\documentclass[journal]{IEEEtran}

\usepackage[utf8]{inputenc}

\DeclareUnicodeCharacter{202F}{ }

\usepackage{amsmath,amsfonts,amssymb}
\usepackage{bm}

\usepackage{graphicx}
\usepackage{booktabs}
\usepackage{multirow}
\usepackage{makecell}
\usepackage{adjustbox}
\usepackage{array}
\usepackage{stfloats}
\usepackage{float}

\usepackage[caption=false,font=normalsize,labelfont=sf,textfont=sf]{subfig}

\usepackage{textcomp}
\usepackage{pifont}

\makeatletter
\DeclareTextFontCommand{\scriptsiz}{\scriptsize}
\makeatother

\usepackage{xcolor}
\usepackage{url}
\usepackage[breaklinks,colorlinks,allcolors=blue]{hyperref}

\usepackage{algorithm}
\usepackage{algorithmic}

\usepackage{cite}

\begin{document}
	
	\title{AgentFoX: LLM-Driven Agentic Multi-Expert Fusion with Explainability for AI-Generated Image Detection}
	
	\author{Yangxin~Yu,~Bin~Li\textsuperscript{*},~Yue~Zhou,%
		~Kaiqing~Lin,~Haodong~Li,~Jiangqun~Ni,~and~Bo~Cao%
		\thanks{Manuscript received \today.}
		\thanks{Y.~Yu, B.~Li, Y.~Zhou, K.~Lin, and H.~Li are with the Guangdong Provincial Key Laboratory of Intelligent Information Processing, Shenzhen Key Laboratory of Media Security, and SZU-AFS Joint Innovation Center for AI Technology, Shenzhen University, Shenzhen 518060, China (e-mail 2453043007@mails.szu.edu.cn; libin@szu.edu.cn).}%
		\thanks{J.~Ni is with the School of Cyber Science and Technology, Sun Yat-sen University, China.}%
		\thanks{B.~Cao is with the Smart City Research Institute of China Electronics Technology Group Corporation, Shenzhen, China.}%
		\thanks{Code \url{https://github.com/suncore946/AgentFoX.git}}%
	}
	
	\markboth{}%
	{Yu \MakeLowercase{\textit{et al.}} AgentFoX LLM-Driven Multi-Expert Fusion for AIGI Detection}
	
	\maketitle
	
	\begin{abstract}
		The realism of AI-generated images (AIGI) poses increasing challenges for reliable forensic detection, where heterogeneous expert detectors may produce conflicting predictions across diverse generative sources and post-processing conditions. 
Existing multi-expert fusion methods rely on fixed rules or learned fusion strategies, offering limited ability to assess sample-specific reliability, execute rigorous adjudication of conflicts, and provide evidence-grounded explanations. We propose AgentFoX, an LLM-driven agentic multi-expert framework for AIGI detection that employs a command-and-reasoning core to perform evidence fusion. 
Following predefined guidelines, the core coordinates designated subtasks to collect semantic and signal-level evidence, reason over structured contexts to determine authenticity, and generate an auditable report for explainability. 
During this process, Expert Profiles are constructed for model-centric reliability assessment, while Clustering Profiles are built for data-centric contextual analysis,  jointly establishing evidence contexts for conflict resolution.
Extensive evaluations across diverse benchmarks demonstrate the robustness and generalizability of AgentFoX under complex conditions.
	\end{abstract}
	
	
	\section{Introduction}
\label{sec:intro}

The rapid advancement of generative AI has produced highly realistic AI-generated images (AIGI) with high visual fidelity and increasingly plausible semantic coherence. 
While these images support applications in entertainment, design, and education, they also introduce serious societal risks, including misinformation and identity forgery, which can further undermine public trust in visual media~\cite{wen2025spot,keita2025deeclip,xu2024fakeshield}.
Ensuring reliable AIGI detection has thus emerged as a critical challenge for multimedia forensics~\cite{yu2024faigc}.

\begin{figure*}[!t]
	\centering
	\subfloat[]{\includegraphics[width=0.58\linewidth, height=4.3cm]{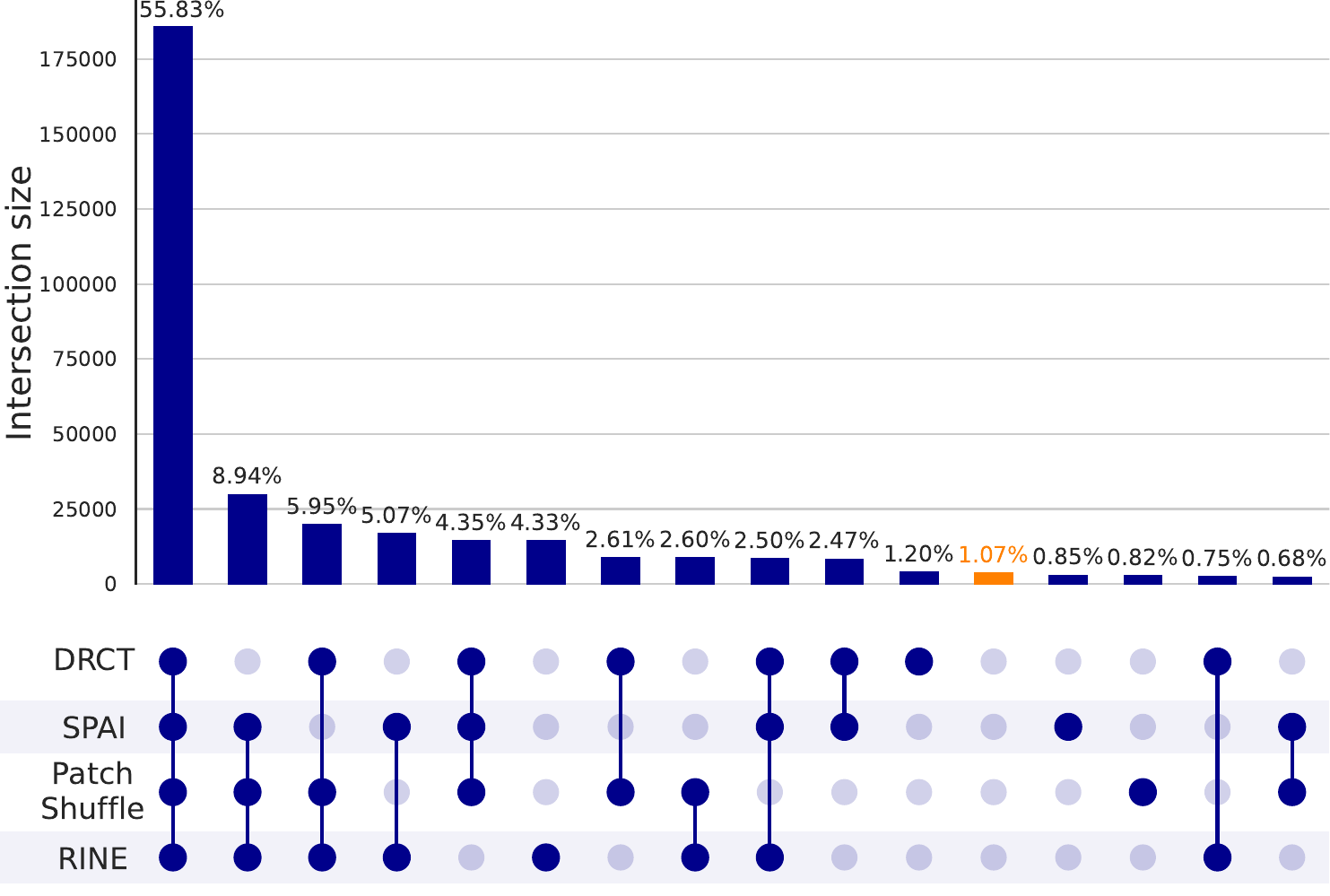}\label{fig:fig1a}}
	\hfil
	\subfloat[]{\includegraphics[width=0.37\linewidth, height=4.3cm]{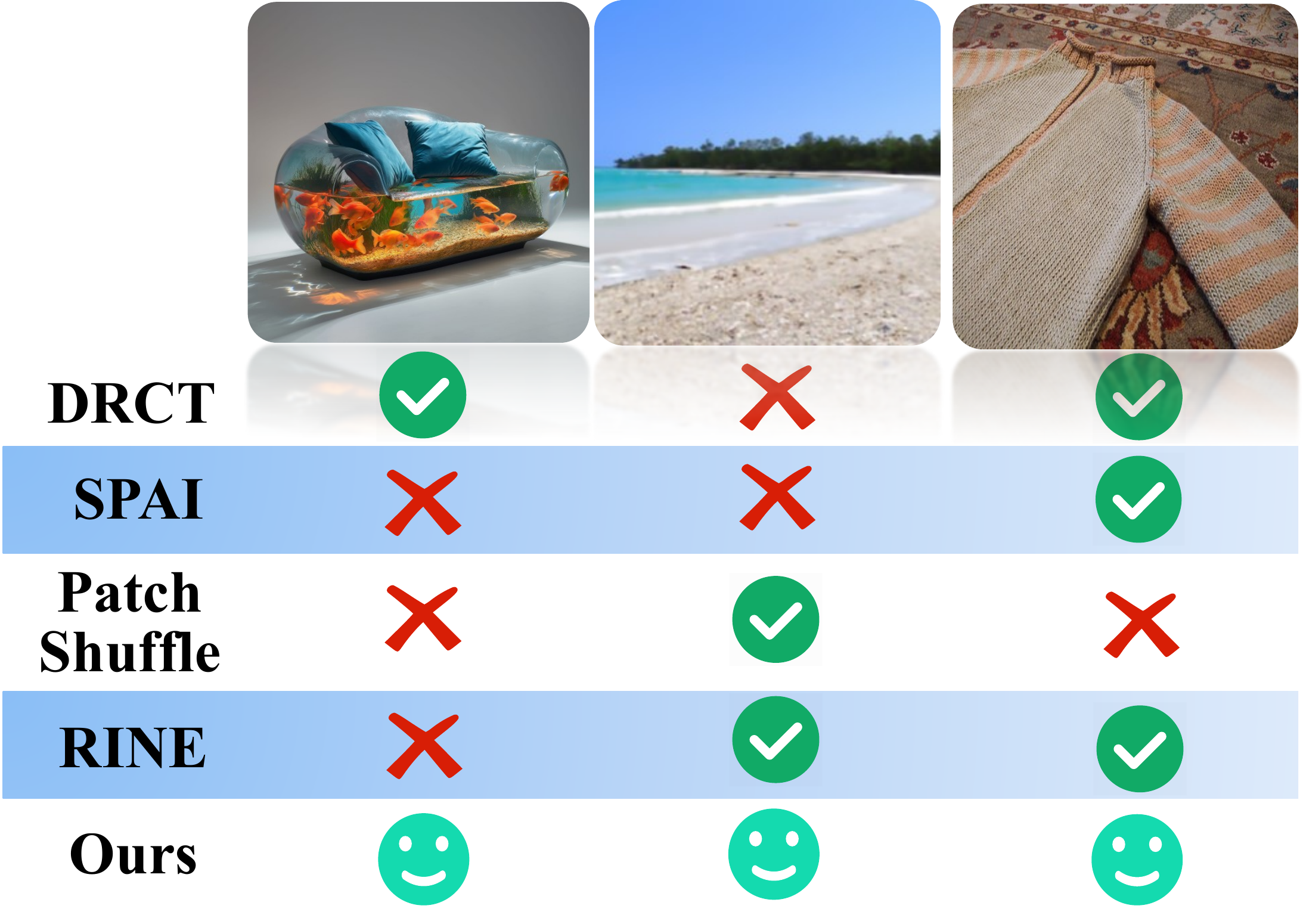}\label{fig:fig1b}}
	\vspace{2mm}
	\caption{\textbf{Consensus and disagreement among AIGI expert detectors.} \textbf{(a)} UpSet plot analysis illustrating complex set intersections of correct predictions among four detectors (DRCT~\cite{Chen2024DRCT}, SPAI~\cite{karageorgiou2025SPAI}, PatchShuffle~\cite{zheng2024breaking}, and RINE~\cite{koutlis2024leveraging}) across seven public datasets(GenImage~\cite{GenImage2023}, AIGCDetect~\cite{zhong2023patchcraft}, AIGIBench~\cite{li2025artificial}, Chameleon~\cite{yan2024sanity}, Community-Forensics~\cite{park2025community}, WIRA~\cite{mcdonald2025real}, and WildFake~\cite{hong2024wildfake}). Filled connected dots indicate the specific subset of experts that correctly classify the exact same samples. The vertical axis denotes the absolute intersection size representing the exact number of samples within each agreement subset, while the percentages above the bars indicate the relative proportion of each subset across the entire evaluated sample pool. \textbf{(b)} Representative high-conflict examples demonstrating scenarios where individual expert predictions contradict one another. The symbols $\checkmark$ and $\times$ denote correct and incorrect expert predictions, while the smiley face indicates that our proposed method successfully resolves the conflict through evidence adjudication and yields the correct decision.}
	\label{fig:fig1}
\end{figure*}

Existing AIGI detectors target specific forensic cues, such as reconstruction artifacts~\cite{Chen2024DRCT}, frequency-domain anomalies~\cite{karageorgiou2025SPAI}, and semantic or commonsense inconsistencies~\cite{koutlis2024leveraging,zheng2024breaking}.
Since different detectors may succeed on distinct types of manipulation evidence, this diversity renders them as complementary sources that collectively capture a broader spectrum of generative artifacts. We refer to these heterogeneous detectors as expert detectors.
However, the reliability of each expert detector may vary across samples, depending on the generative source and training data distribution. 
Consequently, the same image may lead to expert disagreement. To illustrate this disagreement, we evaluate the heterogeneous expert detector set on samples drawn from seven commonly used AIGI evaluation datasets. As shown in Figure~\ref{fig:fig1}, the experts produce inconsistent predictions on 43.1\% of the samples, indicating that expert disagreement is frequent in AIGI detection.
This large proportion of disagreement highlights a key challenge beyond improving individual detectors.
When experts disagree, how should their evidence be fused?

Traditional ensemble strategies~\cite{sagi2018ensemble} typically resolve expert disagreement through aggregation-based fusion. 
Simple rules, such as averaging and majority voting, implicitly assume static expert reliability and are therefore insensitive to context-dependent variations in detector behavior. 
While learned meta-classifiers and gated fusion models offer more reliable fusion, they may require substantial training data to generalize.
They provide limited insight into why certain experts should be trusted for a given image. 
In addition, their fusion mechanisms are usually tied to a fixed set of expert detectors, making the addition or removal of experts burdensome.

These limitations motivate a paradigm shift from aggregation-based approaches to \textit{context-aware evidence fusion}. This paradigm is inspired by human forensic analysis to establish reliability contexts for expert detectors through empirical pattern profiling, and support evidence adjudication.
In implementing this paradigm, we propose \textbf{AgentFoX}, an LLM-driven agentic framework for AIGI detection. AgentFoX employs a \textbf{Command-and-Reasoning Core} to perform context-aware evidence fusion through a structured reasoning workflow where expert prediction is not treated as a definitive conclusion but as a raw evidence source requiring adjudication. The core executes the comprehensive evidence fusion process by invoking a predefined set of \textbf{Forensic Subtasks} encompassing semantic analysis, prior investigation, profiles generation, evidence adjudication, and forensic report formulation. Specifically, the command-and-reasoning core first executes the semantic analysis subtask to extract visual and physical cues. 
It then commands the profile generation subtasks to retrieve relevant records from memory files and integrate them with query samples to generate \textit{Expert Profiles} and \textit{Clustering Profiles}. 
These two profiles focus on describing the signal-level evidence from the perspectives of the model and the data, respectively. 
The memory files are generated by prior investigation tasks, which aim to archive the historical reliability contexts associated with expert detectors, and they can be performed in an off-line manner and reusable for different query samples.
Upon gathering all necessary observations, the Command-and-Reasoning Core executes guideline-regulated forensic reasoning to adjudicate conflicts among heterogeneous detectors. Finally, it consolidates this adjudication process into a structured forensic report to yield an explainable final decision.

Our contributions are summarized as follows:

\begin{itemize}
\item We propose AgentFoX, an LLM-driven agentic framework for multi-expert evidence fusion in AIGI detection. Anchored by a unified \textit{Command-and-Reasoning Core} to coordinate predefined \textit{Forensic Subtasks}, the framework derives multi-faceted evidence from heterogeneous detectors and adjudicates possible conflicts without requiring detector retraining or auxiliary meta-classifiers.

\item We propose two types of textual profiles to encapsulate signal-level detector evidence and facilitate downstream evidence adjudication. \textit{Expert Profiles} adopt a model-centric perspective, describing each detector's intrinsic performance and behavioral characteristics. \textit{Clustering Profiles} capture data-centric reliability by evaluating the detector’s performance on the reference data cluster, which the query sample matches, within the forensic feature space.

\item We systematically evaluate AgentFoX under in-the-wild cross-dataset settings, high-conflict scenarios, and test robustness against image post-processing perturbations. Across these diverse settings, AgentFoX achieves strong performance against individual detectors and fusion baselines, while its forensic reports provide auditable evidence and explainable forensic decisions.
\end{itemize}
	\section{Related Work}
\label{sec:related_works}
\subsection{Signal-level Expert Detectors}
Early efforts in AIGI detection primarily focused on learning low-level signal traces~\cite{wang2020cnn,corvi2023detection} and exploiting statistical anomalies historically derived from conventional digital forensics~\cite{fridrich2012rich, ferrara2012image, tan2024rethinking,tan2024frequencyaware}. As generative models became more capable of suppressing low-level artifacts, later studies shifted to other cues, such as spatiotemporal inconsistencies~\cite{zhang2024deepfake}, disentangled representation learning~\cite{sheng2024dirloc}, and intermediate foundation-model representations~\cite{koutlis2024leveraging}. With the recent prevalence of diffusion models, reconstruction-based methods have emerged to expose diffusion-related artifacts by measuring residual discrepancies between input images and their reconstructed counterparts~\cite{wang2023dire,yao2025reconstruction,ricker2024aeroblade}. Extensive research has also focused on improving detectors through stronger visual backbones~\cite{zhang2025towards,liu2024forgery}, efficient adapter-based tuning for zero-shot generalization~\cite{tan2025c2p,lin2025standing,yan2025orthogonal,Shi2025Transformer,Tang2025Category}, auxiliary supervision strategies~\cite{chen2025dual}, and advanced learning paradigms like soft contrastive learning~\cite{wu2025rethinking}. Furthermore, research on detecting malicious concepts without image generation~\cite{xu2026detecting} and rigorous efforts toward benchmarking and evaluating deepfake detection~\cite{deng2024towards,li2025bridging} provide critical foundations for addressing current forensic challenges.

The differences in model architectures, detection objectives, and training data among signal-level detectors make them respond differently to diverse image sources and generators due to their heterogeneous sensitivities to different forensic traces, often producing conflicting predictions. This motivates our work to resolve conflicts among heterogeneous expert detectors while leveraging their complementary strengths.

\subsection{Semantic-level Reasoning MLLM}
Beyond signal-level detectors, recent studies have adopted Multimodal Large Language Models (MLLMs) to reformulate AIGI detection as a multimodal instruction-following and semantic reasoning task. Existing methods fine-tune MLLMs on multi-modal datasets~\cite{zhang2025ivy, li2025fakescope,huang2025sida,chen2026x2,ji2025interpretable,lin2025seeing} or incorporate external visual knowledge into MLLM-based detection~\cite{li2024fakebench,zhou2025aigi}. These methods make AIGI detection more interpretable, but MLLMs remain limited in perceiving subtle signal-level artifacts~\cite{koutlis2024leveraging,karageorgiou2025SPAI,zheng2024breaking,Chen2024DRCT}. Moreover, their reasoning process does not attempt to reconcile heterogeneous forensic evidence~\cite{yu2025unlocking}.

Recent studies have explored agentic frameworks for visual forensic tasks~\cite{lai2025agent4faceforgery, huang2026unishield}. However, rather than supporting multi-expert evidence fusion, these agents mainly serve as procedural executors in sequential pipelines and do not aim to address the AIGI forensic task.

\subsection{Multi-Expert Fusion and Ensembles}
In AIGI forensics, existing fusion strategies generally follow two paradigms~\cite{cheng2025cospy}: score-level aggregation and learned meta-classifiers. However, both paradigms face notable limitations under high-conflict conditions. Fixed score-level aggregation combines detector outputs in a context-agnostic manner, making it vulnerable to overconfident yet miscalibrated experts. In contrast, trainable meta-classifiers often require costly retraining whenever new experts are introduced, which substantially limits their scalability. Moreover, such models typically function as black boxes, offering limited insight into the historical reliability or evidence patterns of individual sources~\cite{schneider2024genxai}. These limitations motivate a shift toward context-aware evidence fusion, where explicit adjudication mechanisms assess expert credibility by jointly considering historical priors and image-specific cues.
	\section{Method}
\label{sec:method}

\begin{figure*}[!b]
	\centering
	\includegraphics[width=\linewidth]{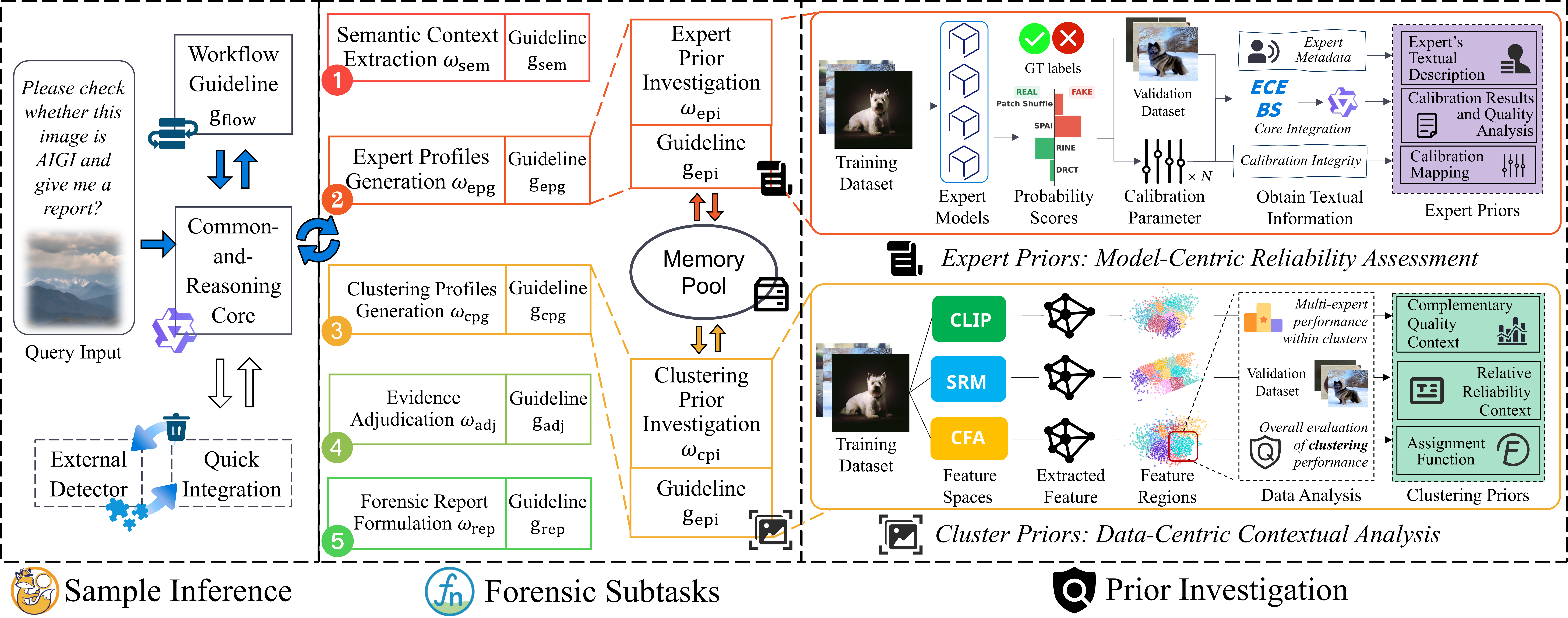}
	\caption{Overview of AgentFoX. 
(Left) During sample inference, the command-and-reasoning core follows predefined workflow guidelines to process queries, invoke tools, and integrate heterogeneous expert detectors.
(Middle) Forensic subtasks provide operational guidelines for constructing agentic memory, establishing historical priors, and linking the memory to the core for profile generation during sample inference.
(Right) Prior investigation constructs expert prior memories via model-centric assessment and cluster prior memories via data-centric analysis.}
	\label{fig:framework_overview}
\end{figure*}

\subsection{Framework Overview}
\label{sec:overview}

In human forensic analysis, practitioners rarely rely on a single cue. Instead, they examine heterogeneous evidence, consider sample-specific conditions, account for the prior reliability of different forensic sources, and adjudicate conflicting observations. However, employing existing agentic frameworks to emulate this evidence fusion process presents inherent limitations. 
On one hand, conventional ReAct-style agentic paradigms rely predominantly on unconstrained generation to drive intermediate reasoning, making them susceptible to logical drift and hallucinated outputs.
In forensic scenarios involving conflicting evidence, unrestricted routing mechanisms may destabilize the reasoning process by triggering recursive tool-use loops or producing inadequately verified conclusions, thereby undermining the reliability of the adjudication.
On the other hand, distributing tasks among multiple agents not only hinders the maintenance of a coherent evidence context but also incurs additional coordination overhead.

Inspired by practitioners' forensic procedures, the proposed AgentFoX framework structurally emulates the human evidence fusion process, as illustrated in Figure~\ref{fig:framework_overview}.
At the center of this framework lies an LLM-based command-and-reasoning core, which coordinates evidence collection and resolves conflicting information through rigorous evidence adjudication.
To achieve this, the overarching workflow decomposes the primary task into a set of subtasks $\Omega$. Each subtask $\omega_{x} \in \Omega$ is strictly governed by a corresponding operational guideline $g_{x} \in \mathcal{G}$, formulated as
\begin{equation}
	\label{eq:framework_sets}
	\begin{aligned}
		\Omega &= \{\omega_{\mathrm{epi}}, \omega_{\mathrm{epg}}, \omega_{\mathrm{cpi}}, \omega_{\mathrm{cpg}}, \omega_{\mathrm{sem}}, \omega_{\mathrm{adj}}, \omega_{\mathrm{rep}}\}, \\
		\mathcal{G} &= \{g_{\mathrm{epi}}, g_{\mathrm{epg}}, g_{\mathrm{cpi}}, g_{\mathrm{cpg}}, g_{\mathrm{sem}}, g_{\mathrm{adj}}, g_{\mathrm{rep}}\}.
	\end{aligned}
\end{equation}
Supported by $\Omega$ and strictly regulated by $\mathcal{G}$, the command-and-reasoning core drives a systematic inference progression. 
This progression encompasses semantic-level evidence perception, signal-level evidence collection, heterogeneous evidence adjudication, and forensic report formulation. 
Such a design ensures that the inference phase maintains modular evidence handling under centralized workflow control, as discussed in Section~\ref{sec:sample_inference}.
To bridge the representational gap between the textual format required by the LLM core and the numerical outputs of expert detectors and agentic tools, signal-level evidence is structured into standardized textual profiles, the benefits of which are described in Section~\ref{sec:Signal-level_Evidence-based_Profiles}.
Furthermore, within the regulatory boundaries of each subtask $\omega_x$ and guided by the corresponding operational guideline $g_x$, the command-and-reasoning core actively issues explicit operational commands.
When needed, the core triggers utilities from a predefined agentic toolkit via these commands to perform targeted operations, such as raw data acquisition, statistical calibration, and memory retrieval.
Facilitating these operations necessitates a command-and-reasoning core capable of structured tool invocation and multi-step logical reasoning.

\subsection{Forensic Subtasks}
\label{sec:sample_inference}

\begin{figure*}[htbp]
	\centering
	\includegraphics[width=\linewidth]{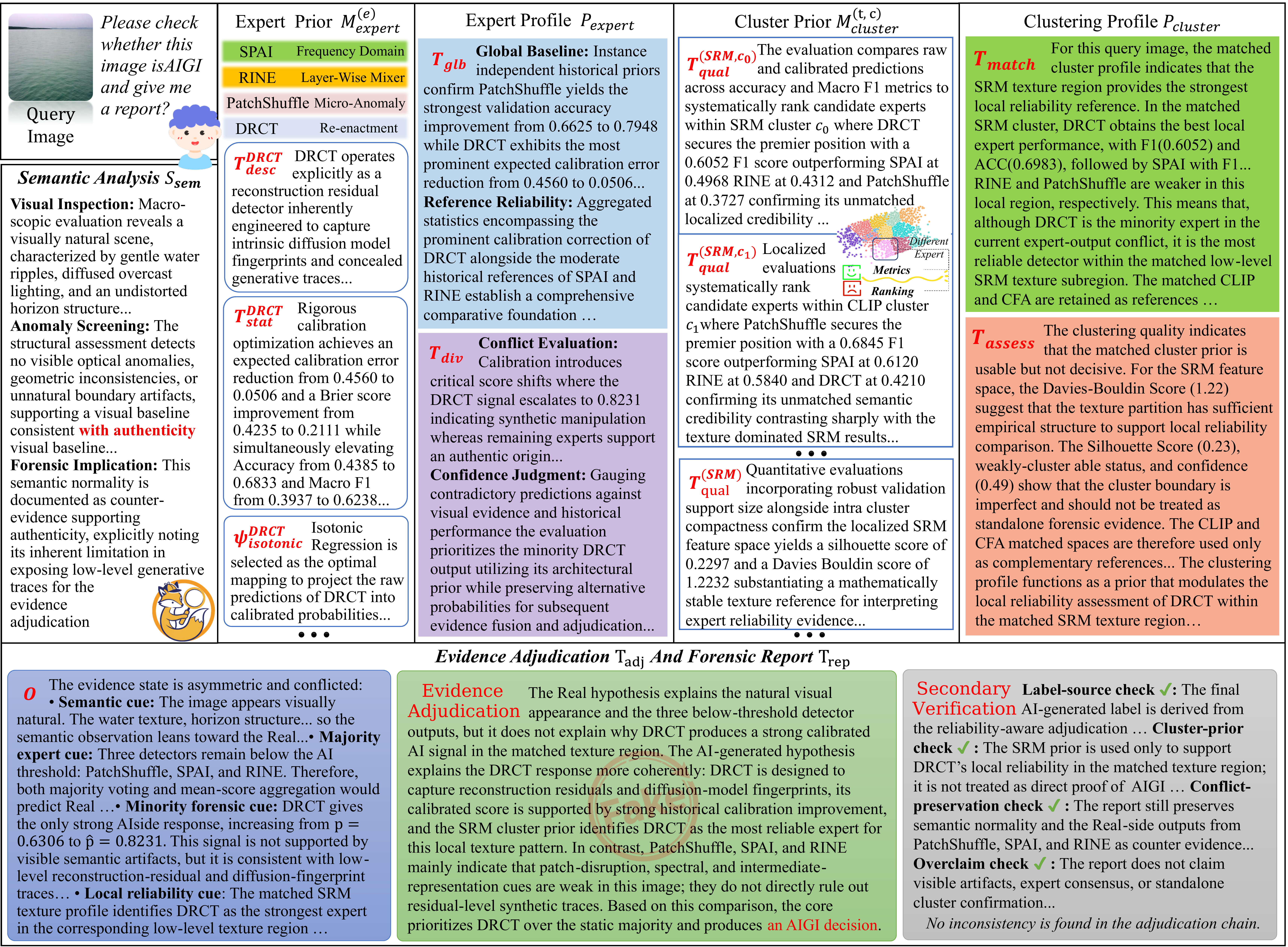}
	\caption{Representative forensic response of AgentFoX. The example illustrates how the framework processes a query image through semantic analysis, signal-level evidence collection, structured evidence adjudication, and forensic report generation.}
	\label{fig:response_case}
\end{figure*}

AgentFoX formally decomposes the AIGI detection workflow into predefined subtasks that enforce explicit execution boundaries governed by specific operational guidelines to prevent reasoning divergence. 
During inference, semantic evidence is first extracted via $\omega_{\mathrm{sem}}$ to capture high-level visual anomalies. 
Subsequently, signal-level evidence is collected hierarchically, with instance-level context generation ($\omega_{\mathrm{epg}}$ and $\omega_{\mathrm{cpg}}$) is conditioned on data-centric historical investigations ($\omega_{\mathrm{epi}}$ and $\omega_{\mathrm{cpi}}$).
The resulting semantic and signal-level outputs are consolidated by $\omega_{\mathrm{adj}}$, where inconsistencies are systematically adjudicated. Finally, the verified rationale is formatted into a structured report by $\omega_{\mathrm{rep}}$. This progression ensures structured and auditable evidence fusion. The overall process is illustrated in Figure~\ref{fig:response_case}.

\subsubsection{Semantic Context Extraction}
\label{sec:semantic_context_extraction}
To capture semantic-level artifacts, the core triggers an external Multi-modal Large Language Model (MLLM) to execute the semantic analysis subtask $\omega_{\mathrm{sem}}$. Guided by $g_{\mathrm{sem}}$, the MLLM analyzes the query image $I$ to detect structural irregularities, geometric inconsistencies, and physically implausible relations. The resulting observations are synthesized into a structured context $\mathcal{S}_{\mathrm{semantic}}$, which serves as independent semantic evidence for subsequent adjudication.
\begin{equation}
	\mathcal{S}_{\mathrm{semantic}} = \omega_{\mathrm{sem}}(I \mid g_{\mathrm{sem}}).
	\label{eq:Semantic_Context}
\end{equation}

\subsubsection{Expert Prior Investigation}
\label{sec:expert_prior_investigation}
Since heterogeneous detectors possess distinct architectural and training priors, their raw outputs lack a unified probabilistic grounding.
To render predictions from heterogeneous detectors comparable for evidence fusion, the core executes the expert prior investigation subtask $\omega_{\mathrm{epi}}$. Following the $g_{\mathrm{epi}}$, the core evaluates and calibrates each candidate detector $E_e$ using a reference dataset $\mathcal{D}$. 
It then synthesizes the empirical evidence into structured memory files, establishing a conceptual tuple for each detector that acts as the foundation for subsequent adjudication.
This tuple is denoted as $\mathcal{M}_{\mathrm{expert}}^{(e)}$ and formulated as
\begin{equation}
	\begin{split}
		\mathcal{M}_{\mathrm{expert}}^{(e)} 
		&= \omega_{\mathrm{epi}} \left( E_e, \mathcal{D} \mid g_{\mathrm{epi}} \right) \\
		&= \left\langle \mathcal{T}_{\mathrm{desc}}^{(e)}, \mathcal{T}_{\mathrm{stat}}^{(e)}, \psi_*^{(e)} \right\rangle, e \in \{1, \dots, N\},
	\end{split}
\end{equation}
where $\mathcal{T}_{\mathrm{desc}}^{(e)}$ contains a general description of expert $E_e$, encompassing its underlying mechanism and typical application scenarios,
$\mathcal{T}_{\mathrm{stat}}^{(e)}$ summarizes calibration-related performance statistics,
and $\psi_*^{(e)}$ denotes the optimal calibration mapping selected from a mapping function set.

The descriptive context $\mathcal{T}_{\mathrm{desc}}^{(e)}$ is formulated by invoking a predefined toolkit to establish the relevant background of $E_e$. Specifically, the core actively extracts this contextual information by analyzing the expert's literature and source codebases.
The statistical summary $\mathcal{T}_{\mathrm{stat}}^{(e)}$ integrates Expected Calibration Error (ECE), Brier Score (BS)~\cite{glenn1950verification}, Accuracy, and Macro-F1 for the expert before and after calibration, thereby providing a robust assessment of both calibration quality and relative classification performance. This multi-metric design mitigates the finite-sample bias introduced by the binning sensitivity of ECE, since BS avoids heuristic binning, while Accuracy and Macro-F1 provide complementary criteria for ranking candidate detectors in terms of classification performance.

Calibration is explicitly defined as a mapping transformation applied to the raw confidence scores produced by $E_e$, thereby rectifying miscalibrated confidence estimates and rendering heterogeneous expert predictions comparable.
Specifically, the reference dataset $\mathcal{D}$ is partitioned into a training subset $\mathcal{D}_{\mathrm{train}}$ and a validation subset $\mathcal{D}_{\mathrm{val}}$. 
The candidate mapping function set $\Psi$ comprises Temperature Scaling~\cite{guo2017calibration}, Platt Scaling~\cite{platt1999probabilistic}, Isotonic Regression~\cite{niculescu2005predicting}, Histogram Binning~\cite{zadrozny2001obtaining}, and Beta Calibration~\cite{kull2017beta}.
These mapping functions are first fitted on $\mathcal{D}_{\mathrm{train}}$ to transform raw scores into well-calibrated probabilities. 
The optimal mapping $\psi^{(e)}_{*}$ is then selected on $\mathcal{D}_{\mathrm{val}}$ by minimizing the ECE~\cite{nixon2019measuring} over validation samples:
\begin{equation}
	\psi_*^{(e)}
	=
	\arg\min_{\psi \in \Psi}
	\mathrm{ECE}\big(\psi(E_e(I));\mathcal{D}_{\mathrm{val}}\big).
	\label{eq:ece_optimization_detailed}
\end{equation}

The subtask $\omega_{\mathrm{epi}}$ can be executed in advance, and its outputs $\mathcal{M}_{\mathrm{expert}}^{(e)}$ can be stored as shared memory files for retrieval by $\omega_{\mathrm{epg}}$. This modular decoupling of the prior memory formulation from the subsequent profile generation process reduces computational overhead and execution latency during sample inference.

\subsubsection{Expert Profiles Generation}
The subtask $\omega_{\mathrm{epg}}$ aims to generate profiles for each expert detector. The profiles are structured textual evidence representations defined by:
\begin{equation}
	\begin{split}
		\mathcal{P}_{\mathrm{expert}}
		&=
		\omega_{\mathrm{epg}}
		\left(
		I,
		\left\{ \mathcal{M}_{\mathrm{expert}}^{(e)} \right\}_{e=1}^{N}
		\mid g_{\mathrm{epg}}
		\right) \\
		&=
		\left\langle
		\mathcal{T}_{\mathrm{glb}},
		\mathcal{T}_{\mathrm{div}}
		\right\rangle,
	\end{split}
	\label{eq:expert_profile_generation}
\end{equation}
where $\mathcal{T}_{\mathrm{glb}}$ delineates the instance-independent confidence baseline derived from historical priors, and $\mathcal{T}_{\mathrm{div}}$ captures the conflicting evidence among the query-specific expert outputs, simultaneously encapsulating a rationale-supported confidence evaluation for each individual expert.

In the subtask $\omega_{\mathrm{epg}}$, each candidate expert $E_e$ is invoked to evaluate the query image $I$, yielding a raw prediction $p^{(e)}$. 
The core then uses the toolkit to retrieve the mapping $\psi_*^{(e)}$ from the prior memory files $\mathcal{M}_{\mathrm{expert}}^{(e)}$ and adjusts the raw prediction into a calibrated prediction $\hat{p}^{(e)}$. 
Constrained by $g_{\mathrm{epg}}$, the core leverages the reasoning capabilities of the LLM to synthesize information from $p^{(e)}$, $\hat{p}^{(e)}$, $\mathcal{T}_{\mathrm{desc}}^{(e)}$, and $\mathcal{T}_{\mathrm{stat}}^{(e)}$ into a unified context. 
Specifically, $\mathcal{T}_{\mathrm{glb}}$ is structurally formulated upon the fixed $\mathcal{D}_{\mathrm{val}}$ and stored priors to encapsulate a comprehensive comparative analysis of the raw and calibrated performance statistics derived from $\{\mathcal{T}_{\mathrm{stat}}^{(e)}\}_{e=1}^{N}$, thereby establishing an entirely instance-independent profile to circumvent repetitive computation during individual image queries.
Conversely, $\mathcal{T}_{\mathrm{div}}$ is structurally formulated to detail the prediction disparities for $I$ across all candidate experts, highlighting contradictions such as inconsistent labels, substantial probability differences, or label changes introduced by calibration, and to contain a core-inferred confidence judgment for each expert, obtained by gauging the expert's prediction against the expert's historical performance.

\subsubsection{Cluster Prior Investigation}
\label{sec:clustering_prior_investigation}

While model-centric priors summarize the overall behavior of the experts, data-cluster priors offer a finer-grained and complementary data-centric perspective by evaluating expertise within data-clustered subregions of the image feature spaces. Governed by the guideline $g_{\mathrm{cpi}}$, the cluster prior investigation subtask $\omega_{\mathrm{cpi}}$ is executed using a reference dataset $\mathcal{D}$ for a set of candidate detectors. The resulting outputs are stored in cluster prior memory files:
\begin{equation}
	\begin{split}
		\mathcal{M}_{\mathrm{cluster}}^{(t, c)}
		&=
		\omega_{\mathrm{cpi}}
		\left(
		\mathcal{D},
		\{E_e, \psi_*^{(e)} \}_{e=1}^{N}
		\mid g_{\mathrm{cpi}}
		\right) \\
		&=
		\left\langle
		\mathcal{T}_{\mathrm{rel}}^{(t,c)},
		\mathcal{T}_{\mathrm{qual}}^{(t)},
		a^{(t)}
		\right\rangle, \\
		&\quad c \in \{1, 2, \dots, K_t\},
		t \in \{1, 2, \dots, T\},
	\end{split}
	\label{eq:clustering_prior_memories}
\end{equation}
where $(t, c)$ denotes the cluster index $c$ in the $t$-th feature space, $T$ represents the total number of considered feature spaces, $\mathcal{T}_{\mathrm{rel}}^{(t, c)}$ records the relative reliability context, $\mathcal{T}_{\mathrm{qual}}^{(t)}$ summarizes the quality context of the clustered feature subregion, and $a^{(t)}$ represents the cluster-assignment function.

This subtask encompasses three key stages. First, it utilizes K-means to cluster image features extracted from the training set $\mathcal{D}_{\mathrm{train}}$, which is inherited from the data split defined in $\omega_{\mathrm{epi}}$. Second, it partitions the validation set $\mathcal{D}_{\mathrm{val}}$ based on the resulting clustered subregions. Third, it evaluates the performance of expert detectors on these partitions and verbalizes the obtained statistical information into the prior memory files in a structured format.

To analyze data in feature spaces with diverse properties, the framework considers a discrete set of $T=3$ complementary representations. Specifically, the three values $t=1,2,3$ correspond to CLIP, SRM, and CFA, respectively, where CLIP characterizes high-level visual-semantic traces, SRM emphasizes low-level noise residuals, and CFA reflects artifacts related to the color filter array.
The core invokes the tool library to project each training sample into the $t$-th feature space via $\mathbf{f}^{(t)}$. Let $\mathcal{Z}_{\mathrm{train}}^{(t)}$ denote the set of extracted training features:
\begin{equation}
	\mathcal{Z}_{\mathrm{train}}^{(t)}
	=
	\left\{
	\mathbf{f}^{(t)}(I_i)
	\mid
	I_i \in \mathcal{D}_{\mathrm{train}}
	\right\}.
\end{equation}
For each feature space $t$, the optimal number of K-means clusters $K_t$ is selected by maximizing the Silhouette score~\cite{rousseeuw1987silhouettes} over a discrete search space $\mathcal{K}$:
\begin{equation}
	K_t
	=
	\arg\max_{K\in\mathcal{K}}
	\mathrm{Sil}
	\left(
	\mathrm{KMeans}_{K}
	\left(
	\mathcal{Z}_{\mathrm{train}}^{(t)}
	\right)
	\right).
\end{equation}

Let $\boldsymbol{\mu}_{\ell}^{(t)}$ for $\ell \in \{1, \dots, K_t\}$ denote the centroid of the $\ell$-th clustered subregion. The assignment function $a^{(t)}(\cdot)$ maps each image sample to its nearest cluster via $\mathbf{f}^{(t)}$, defined as
\begin{equation}
	a^{(t)}(I_i)
	=
	\arg\min_{\ell\in\{1,\dots,K_t\}}
	\left\|
	\mathbf{f}^{(t)}(I_i)
	-
	\boldsymbol{\mu}_{\ell}^{(t)}
	\right\|_2.
\end{equation}
For each feature space $t$, the validation set $\mathcal{D}_{\mathrm{val}}$ is partitioned into cluster-specific subsets as follows:
\begin{equation}
	\mathcal{D}_{\mathrm{val}}^{(t,c)}
	=
	\left\{
	(I_i,y_i)\in\mathcal{D}_{\mathrm{val}}
	\mid
	a^{(t)}(I_i)=c
	\right\}.
\end{equation}

Given the subset $\mathcal{D}_{\mathrm{val}}^{(t,c)}$, the relative reliability context $\mathcal{T}_{\mathrm{rel}}^{(t, c)}$ is generated to systematically record the performance of the raw expert detectors $E_e$ and their calibrated counterparts $\psi_*^{(e)}(E_e)$ on $\mathcal{D}_{\mathrm{val}}^{(t,c)}$ in terms of Accuracy and Macro-F1 scores. 
The core then generates a textual reliability description by ranking the experts according to these metrics. In addition, the core formulates the quality context $\mathcal{T}_{\mathrm{qual}}^{(t)}$ to characterize whether the feature subregion provides a stable reference for interpreting the reliability evidence by summarizing validation support size, intra-cluster compactness, and the Davies-Bouldin score~\cite{davies1979cluster}. Ultimately, this foundational assessment supplies the precise empirical context required for the subsequent evidence adjudication.

\subsubsection{Clustering Profiles Generation}
\label{sec:clustering_profile_generation}
The core executes the clustering profile generation subtask $\omega_{\mathrm{cpg}}$ governed by the guideline $g_{\mathrm{cpg}}$ to establish a query-specific alignment between data-centric priors and instance-level observations for $I$. The clustering profile $\mathcal{P}_{\mathrm{cluster}}$ is formally defined as
\begin{equation}
	\begin{split}
		\mathcal{P}_{\mathrm{cluster}}
		&= \omega_{\mathrm{cpg}} \left( I,
		\left\{ \left\{ \mathcal{M}_{\mathrm{cluster}}^{(t, c)} \right\}_{c=1}^{K_t} \right\}_{t=1}^{T} \mid g_{\mathrm{cpg}} \right) \\
		&= \left\langle \mathcal{T}_{\mathrm{match}}, \mathcal{T}_{\mathrm{assess}} 
		\right\rangle,
	\end{split}
\end{equation}
where $\mathcal{T}_{\mathrm{match}}$ summarizes the historical performance of each expert within the optimally matched cluster $(t,c^*)$ according to the retrieved relational context $\mathcal{T}_{\mathrm{rel}}^{(t,c^*)}$, and $\mathcal{T}_{\mathrm{assess}}$ describes the clustering quality of the matched feature space based on $\mathcal{T}_{\mathrm{qual}}^{(t)}$.

To construct this profile, the core first invokes the assignment function $a^{(t)}$ from the cluster prior memory files to map the query image $I$ to a specific cluster index $c^*$ within each independent feature space $t$ and then retrieves the corresponding contexts from $\mathcal{M}_{\mathrm{cluster}}^{(t, c^*)}$.
Based on the retrieved contexts $\mathcal{T}_{\mathrm{rel}}^{(t, c^*)}$, $\mathcal{T}_{\mathrm{match}}$ is formed to describe the expert performance within the matched data-clustered subregions associated with $I$ across the feature spaces.
Furthermore, $\mathcal{T}_{\mathrm{assess}}$ is formulated based on $\mathcal{T}_{\mathrm{qual}}^{(t)}$ to depict the global feature space separability through the Davies-Bouldin score~\cite{davies1979cluster} and to reflect the local reliability of expert detections within $(t, c^*)$ assessed by intra-cluster compactness and empirical support size.

\subsubsection{Evidence Adjudication}
\label{sec:evidence_adjudication}
To resolve predictive conflicts among heterogeneous experts and semantic observations, the core initiates the evidence adjudication subtask $\omega_{\mathrm{adj}}$. With all contextual evidence prepared, the intermediate outputs encompassing the semantic observations $\mathcal{S}_{\mathrm{semantic}}$, the expert-level profiles $\mathcal{P}_{\mathrm{expert}}$, and the cluster-level profiles $\mathcal{P}_{\mathrm{cluster}}$ are unified into a comprehensive evidence state $\mathcal{O}$ defined by
\begin{equation}
	\begin{aligned}
		\mathcal{O} = \Big\{ 
		& 
		\mathcal{S}_{\mathrm{semantic}},
		\mathcal{P}_{\mathrm{expert}}, 
		\mathcal{P}_{\mathrm{cluster}}		
		\Big\}.
	\end{aligned}
	\label{eq:evidence_state}
\end{equation}
Governed by the adjudication guideline $g_{\mathrm{adj}}$, the core invokes the evidence adjudication subtask $\omega_{\mathrm{adj}}$ to perform logical inference over the evidence state $\mathcal{O}$, producing the final adjudication result:
\begin{equation}
	\mathcal{T}_{\mathrm{adj}}
	=
	\omega_{\mathrm{adj}}
	\left(
	\mathcal{O}
	\mid g_{\mathrm{adj}}
	\right).
\end{equation}
Specifically, the core jointly evaluates the semantic evidence $\mathcal{S}_{\mathrm{semantic}}$ alongside the expert divergence profile $\mathcal{T}_{\mathrm{div}}$, the global statistical context $\mathcal{T}_{\mathrm{glb}}$, and the localized reliability and quality contexts encoded by $\mathcal{T}_{\mathrm{match}}$ and $\mathcal{T}_{\mathrm{assess}}$. Using these multidimensional reliability cues, the core dynamically adjusts the estimated trustworthiness of each expert, discounts uncorroborated outliers, and applies logical reasoning to resolve underlying conflicts, ultimately yielding a unified forensic rationale.

\subsubsection{Forensic Report Formulation}
\label{sec:forensic_report_formulation}
To ensure factual consistency between the preliminary adjudication and the established evidence state, the final reporting subtask $\omega_{\mathrm{rep}}$ is executed. Guided by the reporting guideline $g_{\mathrm{rep}}$, the core employs logical reasoning to reconfirm the adjudication result $\mathcal{T}_{\mathrm{adj}}$ against the observations in $\mathcal{O}$, formulated as
\begin{equation}
	\mathcal{T}_{\mathrm{rep}}
	=
	\omega_{\mathrm{rep}}
	\left(
	\mathcal{T}_{\mathrm{adj}},
	\mathcal{O}
	\mid g_{\mathrm{rep}}
	\right).
\end{equation}
The core explicitly verifies whether the confidence justification and cited forensic cues in $\mathcal{T}_{\mathrm{adj}}$ are supported by the empirical observations recorded in $\mathcal{O}$. If a factual discrepancy is detected, the core flags the inconsistency and reconstructs the 
rationale through evidence-grounded logical inference. 
The resulting $\mathcal{T}_{\mathrm{rep}}$ is an auditable forensic report with verified reasoning.

\subsection{The Advantages of Evidence-Based Profiles}
\label{sec:Signal-level_Evidence-based_Profiles}
The core innovation of the AgentFoX framework is its dual-perspective profiling mechanism, which converts heterogeneous expert evidence into structured textual representations. 
Rather than directly aggregating raw numerical outputs, AgentFoX constructs two complementary profiles: the model-centric expert profile $\mathcal{P}_{\mathrm{expert}}$ and the data-centric clustering profile $\mathcal{P}_{\mathrm{cluster}}$.  
These profiles verbalize each piece of statistical evidence together with its sources, applicable scopes, and uncertainties, enabling the core to interpret expert outputs within an explicit semantic context. 
The dual-perspective textual formulation therefore establishes a structured and interpretable evidence basis for the subsequent subtask $\omega_{\mathrm{adj}}$.

From a model-centric perspective, $\mathcal{P}_{\mathrm{expert}}$ addresses the incompatibility of raw confidence scores produced by heterogeneous experts. 
Because different experts may have different architectures, calibration behaviors, and output distributions, their predictions cannot be compared reliably using raw scores alone. 
$\mathcal{P}_{\mathrm{expert}}$ records historical calibration information and global performance statistics for each expert. 
This profile allows the core to interpret an expert prediction according to its observed statistical reliability rather than treating all confidence scores as perfectly calibrated.

From a data-centric perspective, $\mathcal{P}_{\mathrm{cluster}}$ provides evidence for the current image by referring to validation samples located in the matched local feature regions. 
For each matched region, the reliability context indicates which experts perform better on similar validation samples, while the quality context describes whether the corresponding validation subset provides a stable reference. This complements $\mathcal{P}_{\mathrm{expert}}$  by adding empirical evidence from samples with image features close to the current query image $I$.

The construction of profiles fundamentally decouples prior construction from query-time reasoning. Because empirical statistics are precomputed and archived in memory files, the inference phase only needs to retrieve and verbalize the relevant historical context. 
This decoupling minimizes computational overhead and accommodates the addition of future candidate experts by merely updating memory entries without retraining the core or the expert detectors.
	\section{Experiments}
In this section, we show experiment results to comprehensively evaluate the performance of the proposed AgentFoX framework. First, we
compare AgentFoX with representative baselines on in-the-wild datasets to provide an overall view of detection performance. Second, we evaluate evidence adjudication under scenarios characterized by high-disagreement expert predictions. Third, we
test robustness against common post-processing operations.

\label{sec:experiments}

\subsection{Experimental Settings}
We deploy Qwen3-32B~\cite{yang2025qwen3} as the command-and-reasoning core. GPT-4o~\cite{openai2024gpt4ocard} is employed as an MLLM-based semantic analyzer to execute the subtask $\omega_{\mathrm{sem}}$, as detailed in Section~\ref{sec:semantic_context_extraction}. The numerical operations required by prior investigation and cluster matching, including calibration, feature extraction, and clustering, are encapsulated into a predefined agentic toolkit. The command-and-reasoning core autonomously executes these procedures by issuing explicit commands to invoke the corresponding tool interfaces. We enforce a decoding temperature of 0 and apply a universal random seed of 42 to guarantee experimental reproducibility. We report Macro-F1 and Accuracy as the primary metrics. To ensure diverse and representative data distributions for constructing the prior memory files, we assemble the reference dataset $\mathcal{D}$ from the GenImage training split~\cite{GenImage2023} utilizing an expert-difficulty-aware stratified sampling strategy detailed in Sections~\ref{sec:expert_prior_investigation} and \ref{sec:clustering_prior_investigation}.

\subsubsection{Detector Baselines}
\label{sec:detector_base}
\begin{figure}[htbp]
	\centering
	\includegraphics[width=0.99\linewidth]{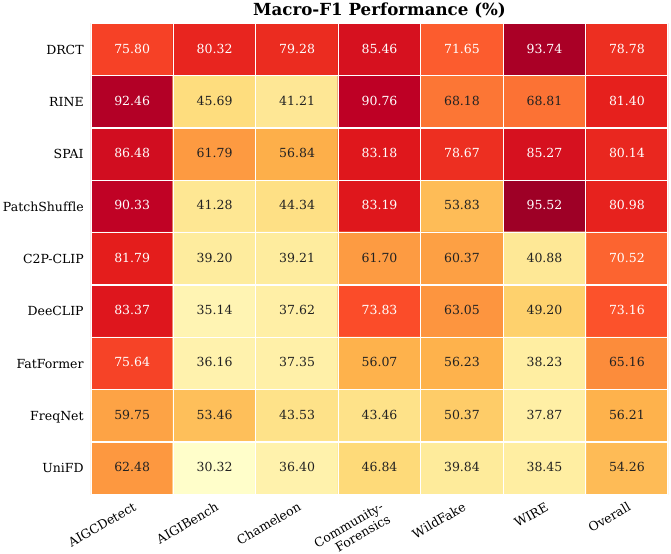}
	\caption{Macro-F1 performance heatmap of the 9 detectors across 6 public AIGI datasets. The dataset-dependent variation reveals complementary strengths and failure modes among DRCT, RINE, SPAI, and PatchShuffle.}
	\label{fig:heatmap}
\end{figure}

We comprehensively evaluate AgentFoX against individual core experts, rule-based and learning-based evidence fusion baselines, signal-level detectors, and semantic-level models. 
To ensure a fair comparison, all core experts and external non-fusion baselines are evaluated via direct inference without any retraining.
AgentFoX instantiates the core expert set as $E_e=\{\mathrm{DRCT}, \mathrm{RINE}, \mathrm{SPAI}, \mathrm{PatchShuffle}\}$. 
We select these four specific models from the nine evaluated baseline detectors because their distinct architectures respectively capture reconstruction residuals, semantic inconsistency, frequency-domain artifacts, and local pattern isolation. 
Figure~\ref{fig:heatmap} demonstrates that these diverse models exhibit highly complementary detection strengths and dataset-dependent failure modes across public AIGI benchmarks. 
The training sources and configurations for all methods are summarized below. We deliberately retain the distinct pre-training configurations of all baseline detectors rather than subjecting them to a uniform retraining protocol. This zero-shot evaluation strategy explicitly preserves the structural heterogeneity required to validate our multi-expert adjudication mechanisms, and simultaneously ensures assessment of cross-domain generalization in practical scenarios.
\begin{itemize}
	\item \textbf{Core Expert Detectors} utilize their publicly released weights. Specifically, RINE~\cite{koutlis2024leveraging} targets Latent Diffusion distributions~\cite{corvi2023detection,rombach2022high}. SPAI~\cite{karageorgiou2025SPAI} relies on self-supervised learning using only real images. PatchShuffle~\cite{zheng2024breaking} and DRCT~\cite{Chen2024DRCT} are trained on the GenImage SDv1.4 subset~\cite{GenImage2023} and the large-scale DRCT-2M dataset~\cite{Chen2024DRCT} respectively.
	
	\item \textbf{Additional Signal-Level Baselines} such as C2P-CLIP~\cite{tan2025c2p}, UniFD~\cite{ojha2023towards}, FatFormer~\cite{liu2024forgery}, FreqNet~\cite{tan2024frequencyaware}, and DeeCLIP~\cite{keita2025deeclip} are primarily trained on ProGAN-based ForenSynths variants~\cite{wang2020cnn}. Certain methods within this category additionally utilize GenImage subsets~\cite{GenImage2023}.
	
	\item \textbf{Semantic-Level Baselines} including FakeVLM~\cite{wen2025spot} and AIGI-Holmes~\cite{zhou2025aigi} are fine-tuned on specialized visual instruction-following datasets, namely FakeClue and Holmes-Set. This supervised training enables these models to leverage high-level semantic and anatomical cues to identify synthetic artifacts. Additionally, we incorporate the foundational multimodal model GPT-4o~\cite{openai2024gpt4ocard} as an independent semantic baseline to establish a reference for zero-shot reasoning performance.
\end{itemize}

\subsubsection{Fusion Baselines}
\label{sec:fusion_baselines}
To compare AgentFoX with alternative fusion algorithms, we include both rule-based and learning-based fusion baselines. 
Rule-based baselines include \textit{Majority Vote} over the four expert detectors and the semantic analyzer~\cite{openai2024gpt4ocard}, and \textit{Probability Average} over the four signal-level expert detectors. 
The semantic analyzer is excluded from probability averaging because it does not provide compatible soft probabilities. 
Learning-based baselines include \textit{MLP Fusion}, \textit{ResNet Fusion}, and \textit{Gated Fusion}, all of which output binary labels using a $0.5$ classification threshold. 
All trainable neural fusion baselines are trained exclusively on the full GenImage train split for 20 epochs, with checkpoint selection based on validation performance. Training strictly on this identically distributed dataset guarantees a fair and comparable evaluation against our framework.

\subsection{In-the-Wild Evaluation}
\label{sec:wild_eval}
To assess out-of-distribution generalization in real-world scenarios, we evaluate different methods on three wild benchmark datasets, namely WIRA~\cite{mcdonald2025real}, WildFake~\cite{hong2024wildfake}, and RRDataset~\cite{li2025bridging}.
Among them, RRDataset further enables evaluation under multiple degradation conditions, including physical recapture, denoted as \textit{Redigital}, and social media transmission, denoted as \textit{Transfer}. 
As shown in Table~\ref{tab:wild_eval}, low-level artifact detectors are competitive on WIRA and RRDataset-Origin, whereas MLLM-based semantic reasoning becomes more reliable under severe physical recapture and social media transmission. Specifically, the standalone GPT-4o baseline demonstrates substantial performance gains on the RRDataset-Transfer task because semantic evidence remains highly informative despite image degradation. However, under social media transmission, GPT-4o occasionally generates cautious semantic analyses accompanied by explicit uncertainty cues. Because the most reliable forensic cue dynamically transitions from pixel-level artifacts to high-level semantics depending on the degradation severity, this evidence shift creates a challenging setting for conventional static fusion strategies. Except for limited gains on WIRA, static baselines such as Majority Vote and Probability Avg, alongside source-trained neural fusion methods including MLP Fusion, ResNet Fusion and Gated Fusion, generally fail to consistently improve upon the strongest individual experts. These results indicate that conventional fusion strategies struggle to match the optimal single expert when the dominant forensic cue changes across target domains.

In contrast, AgentFoX achieves superior performance on WIRA, WildFake and RRDataset-Origin while maintaining high competitiveness on the more severely degraded RRDataset-Redigital and RRDataset-Transfer subsets. Rather than being bottlenecked by the cautious uncertainty cues of the semantic analyzer, the framework leverages them as contextual signals during evidence fusion. The contribution of semantic evidence within AgentFoX depends entirely on how it interacts with the signal-level profiles during the final adjudication process. By adaptively assigning reliability weights based on these interactions, AgentFoX successfully mitigates the limitations of standalone semantic analysis and delivers peak overall performance across all pooled subsets.

\begin{table*}[htbp]
	\centering
	\caption{Generalization and robustness evaluation on in-the-wild benchmarks. Values are reported as Macro-F1 / Accuracy, with the best and second-best results marked in \textbf{bold} and \underline{underline}, respectively. The Overall column pools all subsets before metric computation. Models marked with $^{*}$ are trainable fusion methods.}
	\label{tab:wild_eval}
	\begin{tabular}{ccccccc}
		\toprule
		Model & WIRA & WildFake & RRDataset-Origin & RRDataset-Redigital & RRDataset-Transfer & Overall \\
		\midrule
		DRCT~\cite{Chen2024DRCT} & 0.9374/0.9374 & 0.7165/0.7180 & 0.7868/0.7908 & 0.4712/0.5474 & 0.5437/0.5811 & 0.6432/0.6639 \\
		RINE~\cite{koutlis2024leveraging} & 0.6881/0.7015 & 0.6818/0.6930 & 0.7644/0.7729 & 0.5324/0.5839 & 0.6474/0.6629 & 0.6564/0.6760 \\
		PatchShuffle~\cite{zheng2024breaking} & 0.9552/0.9552 & 0.5383/0.5881 & 0.8559/0.8579 & 0.3783/0.4753 & 0.3354/0.5005 & 0.5951/0.6346 \\
		SPAI~\cite{karageorgiou2025SPAI} & 0.8527/0.8529 & 0.7867/0.7889 & 0.7256/0.7260 & 0.5319/0.5403 & 0.6181/0.6272 & 0.6501/0.6530 \\
		\midrule
		Majority Vote & 0.9060/0.9060 & 0.6044/0.6247 & 0.7563/0.7641 & 0.3732/0.4801 & 0.4063/0.5074 & 0.5672/0.6082 \\
		Probability Avg & \underline{0.9565}/\underline{0.9565} & 0.7903/0.7916 & 0.8620/0.8634 & 0.4767/0.5537 & 0.5555/0.6089 & 0.6800/0.6997 \\
		MLP Fusion$^{*}$ & 0.8627/0.8637 & 0.6381/0.6604 & 0.7563/0.7666 & 0.4239/0.5085 & 0.3478/0.5039 & 0.5657/0.6148 \\	
		ResNet Fusion$^{*}$ & 0.8621/0.8633 & 0.6236/0.6493 & 0.7444/0.7568 & 0.3992/0.5106 & 0.3462/0.5033 & 0.5558/0.6118 \\
		Gated Fusion$^{*}$ & 0.8751/0.8756 & 0.6424/0.6635 & 0.7668/0.7746 & 0.4268/0.5183 & 0.3527/0.5048 & 0.5747/0.6212 \\
		\midrule
		UniFD~\cite{ojha2023towards} & 0.3502/0.4722 & 0.3809/0.4332 & 0.4371/0.5432 & 0.4326/0.5300 & 0.3994/0.5228 & 0.4227/0.5310 \\
		FatFormer~\cite{liu2024forgery} & 0.3880/0.5444 & 0.5301/0.5460 & 0.4852/0.5714 & 0.4350/0.5199 & 0.3382/0.5005 & 0.4246/0.5308 \\
		FreqNet~\cite{tan2024frequencyaware} & 0.4159/0.4306 & 0.4866/0.4866 & 0.5425/0.5489 & 0.5401/0.5403 & 0.4081/0.4613 & 0.5076/0.5160 \\
		C2P-CLIP~\cite{tan2025c2p} & 0.4569/0.5972 & 0.5436/0.5549 & 0.4918/0.5738 & 0.4726/0.5415 & 0.3636/0.5124 & 0.4482/0.5430 \\
		DeeCLIP~\cite{keita2025deeclip} & 0.4624/0.5472 & 0.6008/0.6053 & 0.5852/0.6312 & 0.3829/0.5120 & 0.3489/0.5008 & 0.4504/0.5483 \\
		\midrule
		GPT-4o~\cite{openai2024gpt4ocard} & 0.5333/0.5274 & 0.9074/0.9090 & \underline{0.9005}/\underline{0.9010} & \textbf{0.6654}/\textbf{0.6880} & \textbf{0.7891}/\textbf{0.7920} & \underline{0.7591}/\underline{0.7635} \\
		FakeVLM~\cite{wen2025spot} & 0.8332/0.8346 & 0.6665/0.7015 & 0.7602/0.7721 & 0.6013/0.6082 & 0.6818/0.6818 & 0.6977/0.6983 \\
		AIGI-Holmes~\cite{zhou2025aigi} & 0.4343/0.4980 & \underline{0.9283}/\underline{0.9289} & 0.7295/0.7415 & 0.5835/0.6049 & 0.5756/0.6196 & 0.6323/0.6543 \\
		\midrule
		\textbf{AgentFoX} & \textbf{0.9579}/\textbf{0.9575} & \textbf{0.9326}/\textbf{0.9313} & \textbf{0.9046}/\textbf{0.9105} & \underline{0.6516}/\underline{0.6488} & \underline{0.7116}/\underline{0.7315} & \textbf{0.7817}/\textbf{0.7839} \\
		\bottomrule
	\end{tabular}
\end{table*}

\begin{table*}[htbp]
	\centering
	\caption{Detection performance on the high-conflict X-Fuse benchmark. Results are reported as Macro-F1 / Accuracy, with the best performance \textbf{bolded} and the second best \underline{underlined}. The Overall column is computed over pooled test samples. Models marked with $^{*}$ are trainable fusion methods.}
	\label{tab:combine_result}
	\begin{tabular}{cccccccc}
		\toprule
		Model & AIGCDetect\cite{zhong2023patchcraft} & AIGIBench\cite{li2025artificial} & Chameleon\cite{yan2024sanity} & Community\cite{park2025community}  & WIRA\cite{mcdonald2025real} & WildFake\cite{hong2024wildfake} & Overall \\
		\midrule
		DRCT\cite{Chen2024DRCT} & 0.4779/0.4787 & 0.5744/0.5755 & 0.4616/0.4626 & 0.5377/0.5499 & 0.6043/0.6056 & 0.5387/0.5401 & 0.5238/0.5238 \\
		RINE\cite{koutlis2024leveraging} & 0.4610/0.4610 & 0.4691/0.4729 & 0.4774/0.4809 & 0.5677/0.5812 & 0.4699/0.4722 & 0.5560/0.5579 & 0.4980/0.4980 \\
		SPAI\cite{karageorgiou2025SPAI} & 0.4796/0.4805 & 0.4803/0.4815 & 0.4706/0.4725 & 0.4903/0.5100 & 0.5552/0.5556 & 0.5740/0.5846 & 0.5069/0.5070 \\
		PatchShuffle\cite{zheng2024breaking} & 0.4645/0.4645 & 0.6177/0.6268 & 0.4237/0.4243 & 0.5525/0.5670 & 0.6689/0.6694 & 0.5281/0.5312 & 0.5286/0.5289 \\
		\midrule
		Majority Vote & 0.5565/0.5621 & 0.6785/0.6785 & 0.5153/0.5241 & 0.6469/0.6553 & 0.6346/0.6500 & 0.7209/0.7211 & 0.6162/0.6175 \\
		Probability Avg & 0.5371/0.5372 & 0.5669/0.5676 & 0.4708/0.4709 & 0.6367/0.6524 & 0.6917/0.6917 & 0.6218/0.6231 & 0.5738/0.5743 \\
		MLP Fusion$^{*}$ & 0.8132/0.8156 & 0.5930/0.6142 & 0.5472/0.6556 & 0.6612/0.6695 & 0.6815/0.7028 & 0.6165/0.6202 & 0.6732/0.6862 \\	
		ResNet Fusion$^{*}$ & \underline{0.8201}/\underline{0.8227} & 0.5836/0.6053 & 0.5368/0.6439 & 0.6599/0.6695 & 0.7198/0.7389 & 0.5891/0.5964 & 0.6720/0.6854 \\
		Gated Fusion$^{*}$ & 0.8135/0.8156 & 0.6110/0.6275 & 0.5639/0.6606 & 0.6554/0.6638 & 0.7213/0.7361 & 0.5970/0.6024 & 0.6796/0.6911 \\
		\midrule
		UniFD~\cite{ojha2023towards} & 0.6283/0.6809 & 0.3605/0.5499 & 0.3667/0.5790 & 0.3676/0.3789 & 0.3502/0.4722 & 0.3809/0.4332 & 0.4419/0.5359 \\
		FatFormer~\cite{liu2024forgery} & 0.7854/0.8014 & 0.4146/0.5755 & 0.4094/0.5973 & 0.4814/0.4815 & 0.3880/0.5444 & 0.5301/0.5460 & 0.5506/0.6092 \\
		FreqNet~\cite{tan2024frequencyaware} & 0.7256/0.7376 & 0.6051/0.6610 & 0.5345/0.6489 & 0.4792/0.5043 & 0.4159/0.4306 & 0.4866/0.4866 & 0.5804/0.5983 \\
		C2P-CLIP\cite{tan2025c2p} & 0.7867/0.8014 & 0.4008/0.5584 & 0.4086/0.5957 & 0.5127/0.5128 & 0.4569/0.5972 & 0.5436/0.5549 & 0.5664/0.6193 \\
		DeeCLIP\cite{keita2025deeclip}& 0.7378/0.7589 & 0.4209/0.5783 & 0.4200/0.6057 & 0.5708/0.5726 & 0.4624/0.5472 & 0.6008/0.6053 & 0.5776/0.6229 \\
		\midrule
		GPT-4o~\cite{openai2024gpt4ocard} & 0.6303/0.7358 & \underline{0.8465}/\underline{0.8448} & \underline{0.6699}/\textbf{0.7754} & 0.7119/0.7265 & 0.4468/0.5806 & 0.8564/0.8602 & 0.6636/0.7339 \\
		FakeVLM~\cite{wen2025spot} & \textbf{0.8812}/\textbf{0.8812} & 0.5595/0.6541 & 0.5951/0.6123 & \underline{0.9259}/\underline{0.9402} & \underline{0.7485}/\underline{0.7500} & 0.6922/0.7507 & \underline{0.7393}/\underline{0.7556} \\
		AIGI-Holmes~\cite{zhou2025aigi} & 0.8020/0.8067 & 0.7694/0.7694 & 0.6241/0.6556 & 0.7409/0.7607 & 0.3842/0.4581 & \textbf{0.8850}/\textbf{0.8902} & 0.7221/0.7239 \\
		\midrule
		\textbf{AgentFoX} & 0.7571/0.7571 & \textbf{0.8799}/\textbf{0.8825} & \textbf{0.7079}/\underline{0.7088} & \textbf{0.8565}/\textbf{0.8803} & \textbf{0.7560}/\textbf{0.7583} & \underline{0.8585}/\underline{0.8665} & \textbf{0.7963}/\textbf{0.7977} \\
		\bottomrule
	\end{tabular}
\end{table*}

\subsection{X-Fuse Stress Assessment}
\label{sec:stress_assessment}
To rigorously evaluate multi-expert adjudication under severe disagreement, we 
propose X-Fuse, a diagnostic benchmark designed to isolate high-conflict 
instances in which heterogeneous experts provide contradictory evidence. 
The benchmark is constructed through stratified sampling based on sample-level 
expert correctness patterns derived from the core expert ensemble detailed in 
Section~\ref{sec:detector_base}, where each pattern records which experts 
correctly or incorrectly classify a given candidate sample.

Specifically, we first collect candidate images from the held-out test splits of six diverse AIGI detection datasets, including AIGCDetect~\cite{zhong2023patchcraft}, AIGIBench~\cite{li2025artificial}, Chameleon~\cite{yan2024sanity}, Community-Forensics~\cite{park2025community}, WIRA~\cite{mcdonald2025real}, and WildFake~\cite{hong2024wildfake}. 
Each candidate image is then evaluated by the core expert detectors, and each detector prediction is compared with the ground-truth label to obtain an expert-level correctness pattern. 
Within each dataset and each ground-truth class, images sharing the same correctness pattern are grouped into the same conflict stratum. 
We then sample from each stratum under a fixed quota, which balances different expert-disagreement configurations and prevents the benchmark from being dominated by easy cases where most experts make the same correct prediction. 

Table~\ref{tab:combine_result} summarizes the detection performance on the high-conflict X-Fuse benchmark across individual expert detectors, conventional fusion strategies, state-of-the-art forensic detectors, multi-modal semantic baselines, and the proposed AgentFoX framework.
As an expected consequence of the conflict-driven sampling principle, all individual core experts experience substantial performance degradation across the benchmark. Furthermore, conventional signal-level baselines struggle to generalize under this severe disagreement, and standalone semantic-level models exhibit extreme performance variance due to their exclusive reliance on macro-level visual knowledge. While trainable neural fusion baselines generally outperform single experts by learning static combinations, they fail to adapt to context-dependent reliability shifts across diverse target domains. AgentFoX delivers the optimal overall Macro-F1 score and Accuracy without relying on full-parameter fine-tuning. This superior stability is achieved because the framework adjudicates conflicts by combining independent semantic observations with structured signal-level reliability profiles.

\begin{figure}[htbp]
	\centering
	\includegraphics[width=1\linewidth]{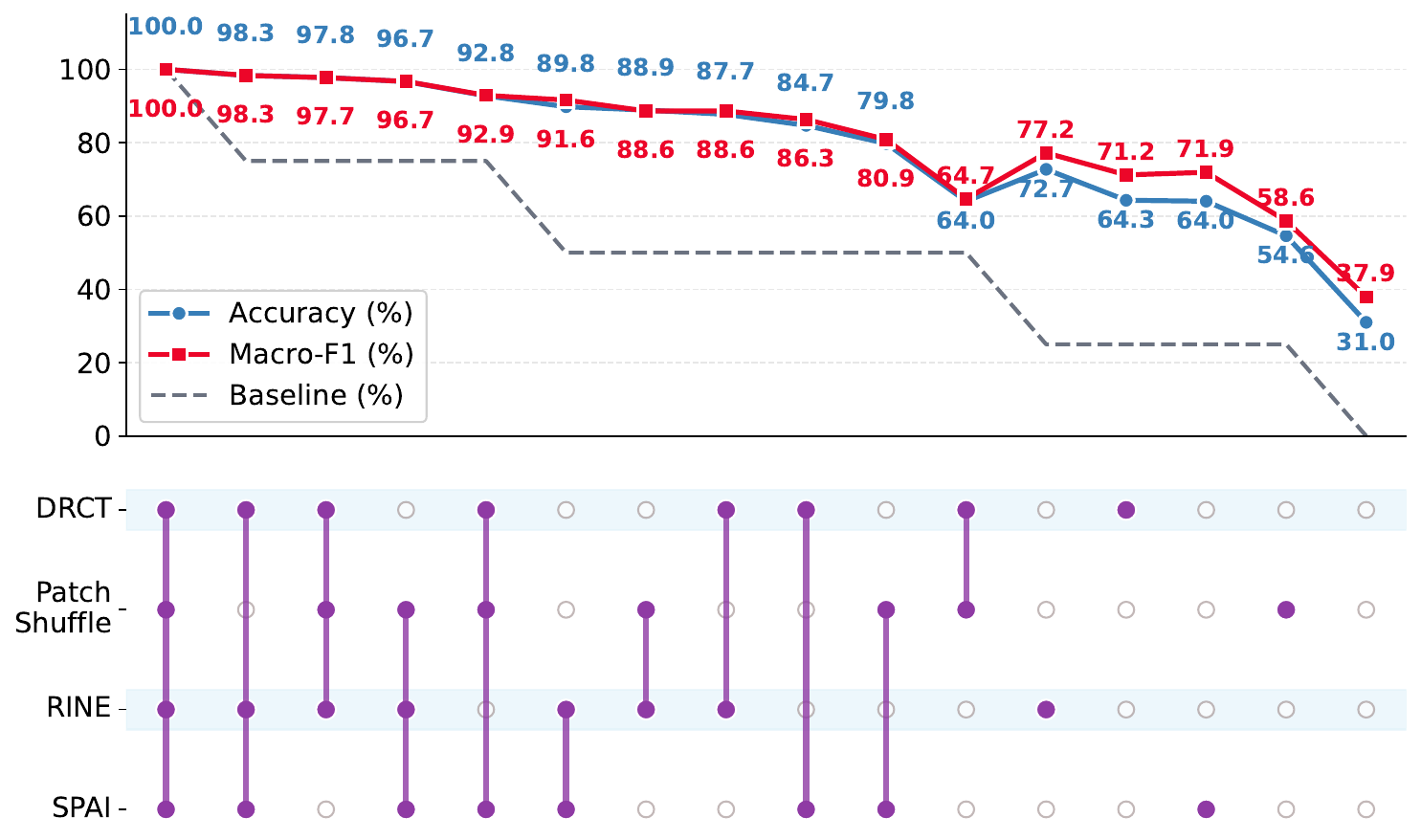}
	\caption{Performance across 16 stratified expert-conflict scenarios, demonstrating the stability of AgentFoX under diverse disagreement conditions. The baseline denotes the average empirical accuracy of all candidate experts, mathematically equivalent to a uniformly random selection.}
	\label{fig:linechart}
\end{figure}

This robustness is examined across the 16 stratified expert-conflict patterns in Figure~\ref{fig:linechart}.
AgentFoX maintains more stable performance than conventional fusion baselines when reliable evidence remains available within the expert panel.
In scenarios where the current expert set lacks reliable evidence for a specific conflict pattern, additional experts can be incorporated by constructing their corresponding prior memories and profiles, without modifying the overall adjudication workflow.

\subsection{Robustness to Multimedia Transmission Artifacts}
\label{sec:robustness_perturbations}
AIGI shared on social platforms often undergoes post-processing that weakens low-level forensic traces.
We evaluate robustness under four perturbations with varying intensities: JPEG compression ($Q \in \{100, 90, 80, 70\}$), isotropic resizing ($s \in \{0.6\times, 0.8\times, 1.2\times, 1.4\times\}$), Gaussian blur ($\sigma \in \{0.5, 1, 2, 3\}$), and center cropping ($512\times512$ to $224\times224$).
All compared detectors use their publicly released checkpoints, without additional retraining, fine-tuning, or perturbation-specific adaptation in our experiments, although their original training protocols may include data augmentation.

\begin{figure*}[!t]
	\centering
	\includegraphics[width=\textwidth]{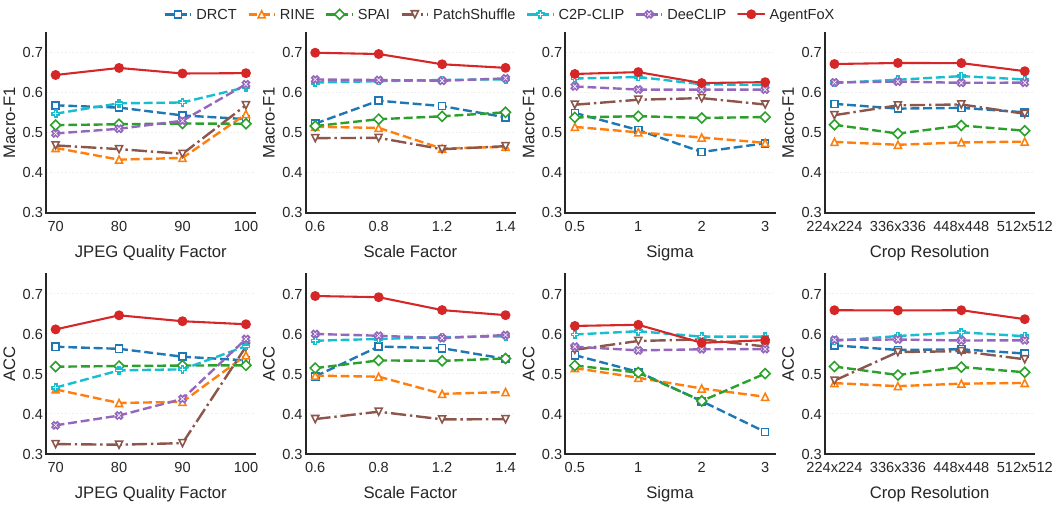}
	\caption{Robustness evaluation under common post-processing perturbations. All evaluations are conducted on the X-Fuse dataset. The top and bottom rows illustrate the Accuracy and Macro-F1 scores, respectively, against JPEG compression, resizing, Gaussian blur, and center crop. AgentFoX demonstrates superior stability compared with the evaluated baselines across various perturbation intensities.}
	\label{fig:robustness_matrix}
\end{figure*}

Figure~\ref{fig:robustness_matrix} shows that AgentFoX maintains stable performance across the tested perturbations and consistently outperforms individual baseline detectors.
This robustness does not come from perturbation-specific retraining or adaptation, but from the reliability-guided adjudication mechanism of AgentFoX.
Specifically, heterogeneous experts provide complementary forensic evidence, profile memories provide reliability context under different feature-space regions and input conditions, and the final evidence fusion adaptively assigns less confidence to experts whose cues become unreliable after post-processing.
As a result, when JPEG compression suppresses high-frequency artifacts, resizing and cropping disturb spatial or scale-dependent traces, or Gaussian blur smooths local texture cues, AgentFoX can still rely on the experts and evidence sources that remain more reliable.
These results demonstrate that AgentFoX improves robustness through adaptive multi-expert evidence fusion rather than perturbation-specific model tuning.

\subsection{Ablation Study}
\label{sec:ablation}

We conduct ablation experiments to examine the contributions of key components in AgentFoX.

\subsubsection{Profile Evidence}

Table~\ref{tab:ablation_profiles} isolates the roles of Expert Profiles and Clustering Profiles.
Expert Profiles provide model-centric reliability evidence for signal-level experts, while Clustering Profiles provide local reliability evidence associated with similar artifact patterns.
The results show that each profile type contributes to performance, and their combination yields the strongest result.
This confirms the benefit of jointly using global expert behavior and local artifact-aware reliability evidence.

\begin{table}[!htbp]
	\centering
	\caption{Ablation of Expert and Clustering Profiles.} 
	\label{tab:ablation_profiles}
	\begin{tabular}{cc|cc}
		\toprule
		\textbf{Expert Profiles} & \textbf{Clustering Profiles} & \textbf{Macro-F1} & \textbf{Accuracy} \\
		\midrule
		$\times$ & $\times$ & 0.7032 & 0.6738 \\
		\checkmark & $\times$ & 0.7177 & 0.6836 \\
		$\times$ & \checkmark & 0.7597 & 0.7376 \\
		\checkmark & \checkmark & \textbf{0.7963} & \textbf{0.7977} \\
		\bottomrule
	\end{tabular}
\end{table}

\subsubsection{Core LLM Backbone}

To assess the dependency on the core adjudication model, we replace the default LLM with alternative backbones of different scales, as shown in Table~\ref{tab:mllm_ablation}.
The lightweight Qwen3-4B model shows a clear performance drop, suggesting that sufficient reasoning capacity is important for multi-expert evidence assessment.
Models at the 8B scale or above recover much of the performance, although results vary across architectures.
The default Qwen3-32B backbone achieves the strongest result among the evaluated alternatives.

\begin{table}[!htbp]
	\centering
	\caption{Ablation study on the core adjudication model dependency. Results denote Macro-F1 and Accuracy on the high-conflict X-Fuse benchmark.}
	\label{tab:mllm_ablation}
	\begin{tabular}{lcc}
		\toprule
		\textbf{Agentic Backbone} & \textbf{Macro-F1} & \textbf{Accuracy} \\
		\midrule
		Qwen3-4B  & 0.4759 & 0.6053 \\
		Qwen3-8B  & 0.7451 & 0.7329 \\
		InternVL3-8B & 0.6900 & 0.7268 \\
		Qwen2.5-VL-32B & 0.7060 & 0.7266 \\ 
		Qwen3-VL-32B & 0.7381 & 0.7840 \\
		\midrule
		\textbf{Qwen3-32B (Ours)} & \textbf{0.7963} & \textbf{0.7977} \\
		\bottomrule
	\end{tabular}
\end{table}

\subsubsection{Semantic Analyzer}
\begin{table}[!htbp]
	\centering
	\caption{Ablation of different semantic analyzers on X-Fuse. Values denote Macro-F1 / Accuracy.}
	\label{tab:ablation_semantic_analyzer}
	\begin{tabular}{lcc}
		\toprule
		\shortstack{\textbf{Semantic}\\\textbf{Analyzer}} 
		& \shortstack{\textbf{Standalone Forensic}\\\textbf{Performance}} 
		& \shortstack{\textbf{AgentFoX w/ Analyzer}\\\textbf{Performance}} \\
		\midrule
		GPT-4o & 0.6636 / 0.7339 & 0.7963 / 0.7977 \\
		GPT-5.4 & 0.7029 / 0.7086 & 0.7655 / 0.7656 \\
		Qwen3-VL-235B & 0.6322 / 0.6537 &  0.6971 / 0.7117 \\
		FakeVLM & 0.7393 / 0.7556 & 0.7373 / 0.7619 \\
		\bottomrule
	\end{tabular}
\end{table}

Table~\ref{tab:ablation_semantic_analyzer} evaluates the architectural dependency of AgentFoX on the chosen semantic analyzer using the high-conflict X-Fuse benchmark. The results show that stronger standalone forensic performance does not necessarily translate into better performance after integration into AgentFoX. For example, GPT-5.4 and FakeVLM achieve higher standalone Macro-F1 than GPT-4o, yet AgentFoX obtains the best overall performance when using GPT-4o as the semantic analyzer. This suggests that the semantic analyzer is not expected to act as an independent forensic classifier, but rather to provide complementary visual and physical context for the evidence adjudication subtask $\omega_{\mathrm{adj}}$.
In particular, FakeVLM is explicitly optimized for forgery detection, which encourages it to produce more verdict-oriented forensic judgments. Such outputs can partially bypass the intended passive contextual role of the semantic analyzer, making it harder for the reasoning core to reconcile semantic descriptions with signal-level expert profiles. In contrast, GPT-4o tends to provide descriptive visual irregularities without enforcing a definitive forensic verdict, allowing the reasoning core to leverage it more effectively as auxiliary evidence. Consequently, effective multi-expert evidence fusion requires the semantic analyzer to supply descriptive physical constraints.

\subsubsection{Expert Integration}
\label{sec:expert_integration}

Given that integrating novel detectors necessitates standardized interface registration and profile construction, Table~\ref{tab:expert_integration} evaluates integration boundaries exclusively within our selected four-expert set. 
Exploring the combinatorial space reveals that configurations covering more diverse forensic cues outperform combinations with overlapping evidence sources.
This performance gap indicates that effective evidence fusion depends not only on the number of experts, but also on whether the expert panel provides sufficiently diverse signal-level evidence.
This confirms that when a specific forensic dimension is absent, the large language model acting as the reasoning core of the agent lacks sufficient complementary signals to execute robust adjudication, thereby establishing the complete heterogeneous ensemble of DRCT, RINE, SPAI, and PatchShuffle as a minimal yet optimally sufficient panel.

\begin{table}[htbp]
	\centering
	\caption{Combinatorial and leave-one-out ablation study on the core expert set, showing that diverse signal-level evidence is more important than merely increasing the number of experts.}
	\label{tab:expert_integration}
	\resizebox{\columnwidth}{!}{
		\begin{tabular}{lcccc|cc}
			\toprule
			\textbf{Strategy} & \textbf{DRCT} & \textbf{RINE} & \textbf{SPAI} & \textbf{PatchShuffle} & \textbf{Macro-F1} & \textbf{Accuracy} \\
			\midrule
			Best Single & & & & \checkmark & 0.5286 & 0.5289 \\
			\midrule
			Redundant Pair & & \checkmark & \checkmark & & 0.4771 & 0.5464 \\
			Orthogonal Pair & \checkmark & & & \checkmark & 0.5317 & 0.5411 \\
			\midrule
			\multirow{4}{*}{Leave-One-Out} 
			& \checkmark & \checkmark & & \checkmark & 0.5365 & 0.6059 \\
			& \checkmark & & \checkmark & \checkmark & 0.5739 & 0.6460 \\
			& \checkmark & \checkmark & \checkmark & & 0.5920 & 0.6511 \\
			& & \checkmark & \checkmark & \checkmark & 0.6331 & 0.6582 \\
			\midrule
			Full Core Experts & \checkmark & \checkmark & \checkmark & \checkmark & \textbf{0.7963} & \textbf{0.7977} \\
			\bottomrule
		\end{tabular}
	}
\end{table}

%


\subsection{Explainability}
\label{sec:explainability}
\begin{figure}[htbp]
	\centering
	\includegraphics[width=0.99\linewidth]{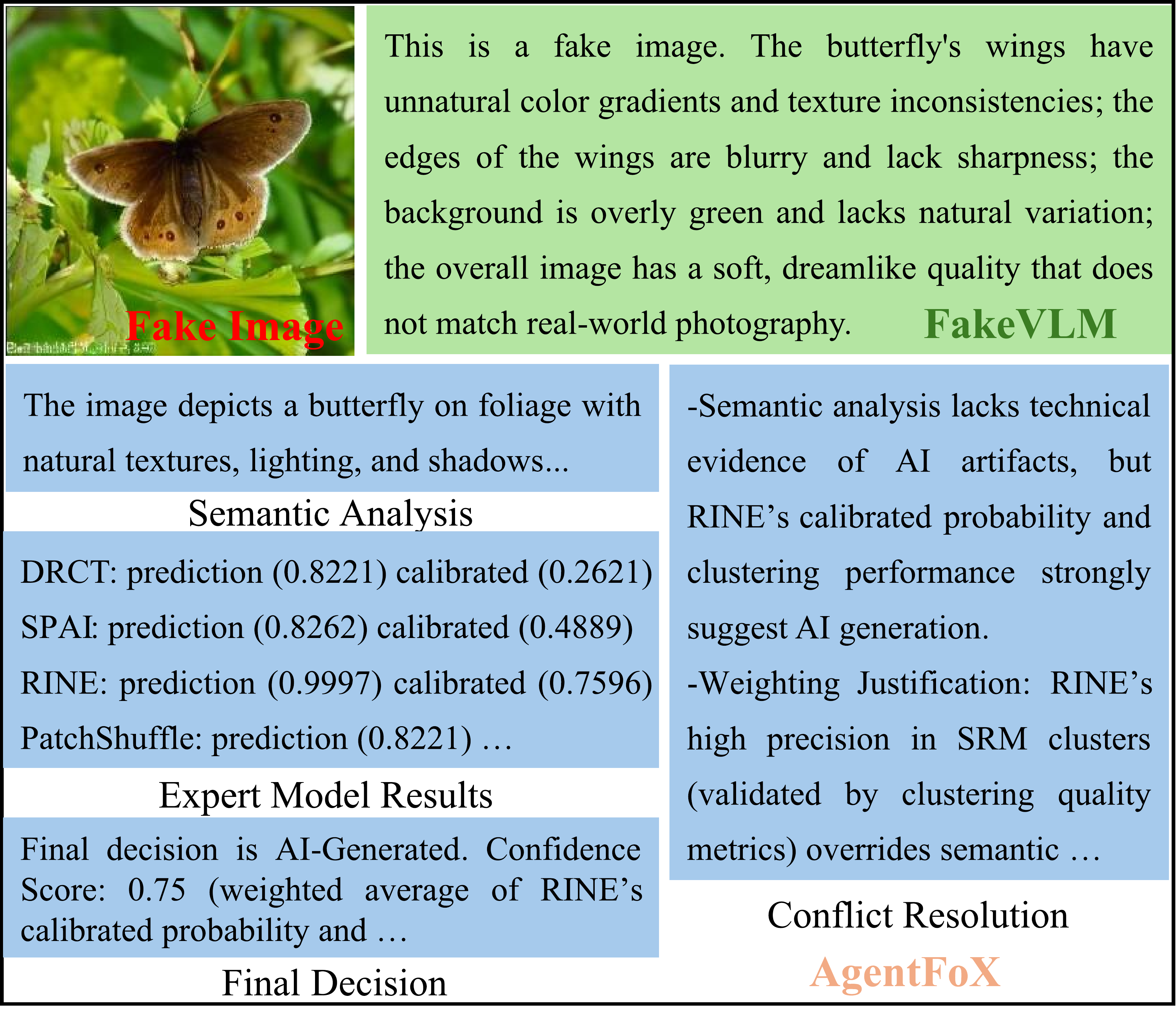}
	\caption{\textbf{Representative explainability comparison between FakeVLM and AgentFoX.} Top: FakeVLM provides qualitative semantic descriptions. Bottom: AgentFoX combines multi-expert signal-level evidence and structured reporting to produce a transparent forensic decision.}
	\label{fig:explainability}
\end{figure}
To illustrate the interpretability of AgentFoX, Figure~\ref{fig:explainability} presents a representative case.
Existing explainable AIGI detectors typically provide either visual saliency, isolated artifact cues, or high-level textual descriptions, which makes it difficult to trace how heterogeneous evidence leads to the final decision.
In contrast, AgentFoX produces a structured forensic report that links multi-expert signal-level evidence, confidence estimates, and final prediction into a unified explanation.
Compared with FakeVLM, which mainly provides high-level semantic descriptions, AgentFoX further includes a quantitative confidence score and an auditable reasoning report.
This makes the decision process more traceable for forensic inspection scenarios.

\subsection{Inference Cost}
\label{sec:cost_tradeoff}

AgentFoX incurs higher inference overhead than direct signal-expert inference and lightweight baselines.
As shown in Table~\ref{tab:cost_appendix}, it processes each image in approximately 27.2 seconds, with about 25.6k total tokens including textualized expert outputs, profile descriptions, prompts, and final report generation.
This overhead mainly comes from multi-expert evidence collection and structured report generation.

\begin{table}[htbp]
	\centering
	\caption{Inference cost and performance on X-Fuse. Latency and token usage are averaged per image. ``Signal Experts'' denotes direct use of signal-level detectors without agent-based reasoning or report generation.}
	\label{tab:cost_appendix}
	\begin{tabular}{lccc}
		\toprule
		\textbf{Metric} & \textbf{Signal Experts} & \textbf{FakeVLM} & \textbf{AgentFoX} \\
		\midrule
		Avg Latency (s) & $<$2 & 7.69 & 27.20 \\
		Avg Input Tokens & N/A & $\sim$580 & $\sim$22000 \\
		Avg Output Tokens & N/A & $\sim$150 & $\sim$3614 \\
		Avg Estimated FLOPs & N/A & 15.05 & 164.35 \\
		Avg Accuracy & 0.5122 & 0.7556 & 0.7977 \\
		\bottomrule
	\end{tabular}
\end{table}

Despite the additional cost, AgentFoX achieves stronger detection performance than direct signal-level expert inference and semantic-level MLLM on X-Fuse.
This trade-off is practical for latency-tolerant forensic scenarios where accuracy, traceability, and structured evidence are prioritized.

\section{Limitations} 
\label{sec:limitations}
\begin{figure}[htbp]
	\centering
	\includegraphics[width=0.99\linewidth]{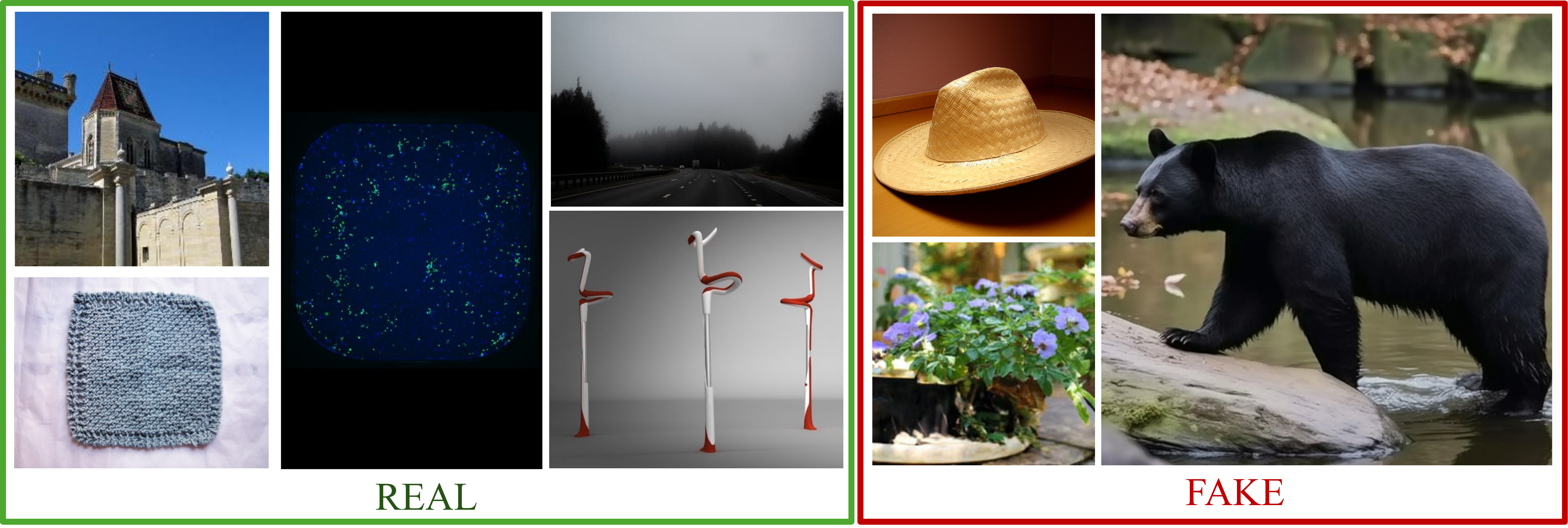}
	\vspace{0.15cm}
	\rule{0.99\linewidth}{0.4pt}
	\vspace{0.15cm}
	\includegraphics[width=0.99\linewidth]{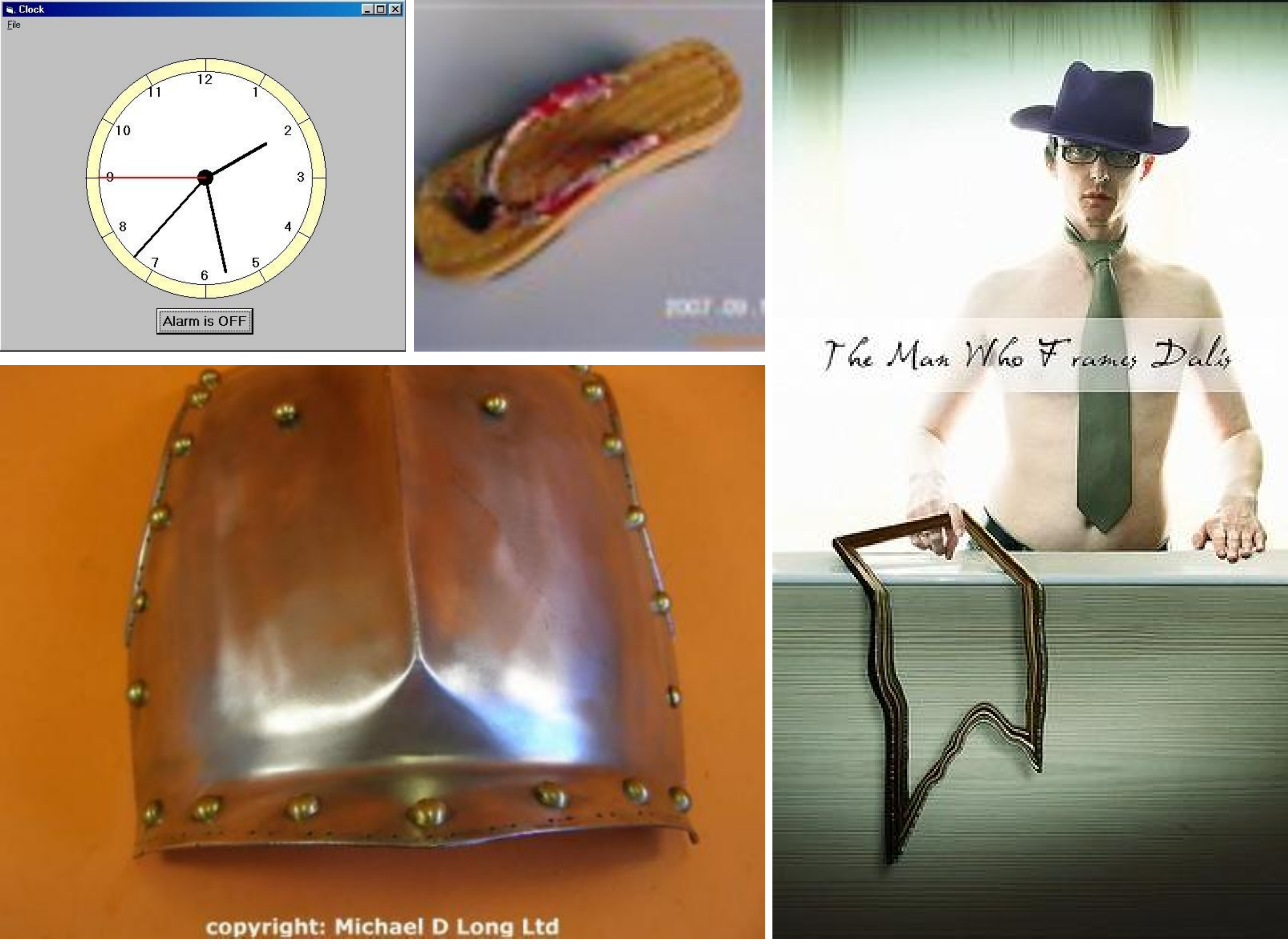}
	\caption{The upper panel illustrates specific error cases driven by reasoning-drift false negatives. The lower panel provides a detailed label analysis of misclassified authentic images featuring screenshots, extreme blur, or digital watermarks to highlight signal-driven false positives.}
	\label{fig:limitations}
\end{figure}

While AgentFoX demonstrates robust adjudication capabilities, systematic analysis of edge cases presented in Figure~\ref{fig:limitations} provides critical insights into the operational boundaries of the current framework.
Signal-driven false positives occur because atypical visual properties inherent to severely degraded or non-photographic inputs systematically compel low-level experts to extract unreliable evidence, demonstrating that overall efficacy of the multi-expert evidence fusion naturally depends on the baseline robustness of foundational detectors.
Moreover, the LLM-based reasoning core currently lacks explicit confidence estimation, making it difficult to distinguish uncertain adjudications from high-confidence decisions under conflicting or degraded expert evidence.
Furthermore, reasoning-drift false negatives emerge during extended multi-stage inference since the LLM reasoning core occasionally allows strong semantic priors to overshadow subtle pixel-level flaws, leading to the misclassification of highly coherent forgeries.
Enhancing robustness of the adjudication process against these complex edge conditions constitutes our primary direction for future research.

%
%
	\section{Conclusion}
\label{sec:conclusion}
In this paper, we present AgentFoX, an LLM-driven agentic framework for multi-expert evidence fusion in AIGI detection. The proposed framework reformulates this task as evidence adjudication over heterogeneous forensic signals, where predefined forensic subtasks and explicit operational guidelines constrain the reasoning process. By separating prior memory construction from sample inference, AgentFoX establishes reusable historical reliability priors and generates Expert Profiles and Clustering Profiles as instance-level textual contexts to ground evidence adjudication. Experiments under in-the-wild settings, high-conflict expert disagreement, and post-processing perturbations show that AgentFoX achieves more robust performance than individual experts and conventional fusion baselines. The framework can also incorporate newly introduced detectors by constructing their corresponding prior memories and profiles, without requiring retraining of the reasoning core or auxiliary meta-classifiers. In addition to improving prediction performance, AgentFoX moves beyond opaque score aggregation by formulating structured forensic reports, thereby improving the traceability and auditability of AIGI forensic analysis. Despite these benefits, AgentFoX remains affected by the representativeness of the reference dataset used for prior memory construction, the absence of calibrated confidence estimates from the Semantic Analyzer, and the latency and token cost introduced by multi-stage LLM inference. Future work will explore adaptive profile updating, confidence-aware semantic analysis, compact textual representations, and efficient evidence adjudication mechanisms that preserve transparency while reducing inference overhead.
	
	{
		\bibliographystyle{IEEEtran}
		\bibliography{IEEEabrv,main}
	}
	
\end{document}